\documentclass[10pt,twocolumn,letterpaper]{article}

\usepackage{cvpr}              

\usepackage[accsupp]{axessibility}  
\usepackage{graphicx}
\usepackage{amsmath}
\usepackage{amssymb, amsfonts}
\usepackage{amsmath,amssymb}
\usepackage{booktabs}
\usepackage{rotating}
\usepackage{array}
\usepackage{multirow}
\usepackage{comment}
\usepackage{diagbox}
\usepackage{epsfig}
\usepackage[ruled, vlined, linesnumbered]{algorithm2e}
\usepackage{subcaption}
\usepackage{amsthm}

\usepackage{threeparttable} 
\usepackage{mathrsfs}
\usepackage{enumitem}
\usepackage{lipsum}
\usepackage{listings}
\usepackage[final]{microtype}
\usepackage{soul}
\usepackage{makecell}
\usepackage[dvipsnames]{xcolor,colortbl}
\usepackage{tabularx,ragged2e}
\usepackage{pifont}
\usepackage{dsfont}
\usepackage{float}

\newcommand{\Devavrat}[1]{\textcolor{MidnightBlue}{#1}} 

\newcommand{\std}[1]{{\scriptsize$\pm$#1}}
\newcommand{\apxA}[0]{\ref{appendix:hyperparameters}}
\newcommand{\apxB}[0]{\ref{appendix:additional_results}}

\newcommand{\apxE}[0]{\ref{appendix:sensitivity}}

\definecolor{backcolour}{rgb}{0.95,0.95,0.92}
\definecolor{codegreen}{rgb}{0,0.6,0}

\lstdefinestyle{myStyle}{
    backgroundcolor=\color{backcolour},   
    commentstyle=\color{codegreen},
    basicstyle=\ttfamily\footnotesize,
    breakatwhitespace=false,         
    breaklines=true,                 
    keepspaces=true,                 
    numbers=left,       
    numbersep=5pt,                  
    showspaces=false,                
    showstringspaces=false,
    showtabs=false,                  
    tabsize=2,
}
\usepackage[pagebackref,breaklinks,colorlinks]{hyperref}

\usepackage[capitalize]{cleveref}
\crefname{section}{Sec.}{Secs.}
\Crefname{section}{Section}{Sections}
\Crefname{table}{Table}{Tables}
\crefname{table}{Tab.}{Tabs.}

\def\cvprPaperID{6724} 
\def\confName{CVPR}
\def\confYear{2023}

\begin{document}

\newcommand{\aug}{\tilde{\boldsymbol{x}}}
\newcommand{\im}{\boldsymbol{x}}
\newcommand{\ps}{\boldsymbol{y}}
\newcommand{\glob}{\bar{\boldsymbol{z}}}
\newcommand{\dense}{\mathbf{z}}
\newcommand{\denseu}{\mathbf{u}}
\newcommand{\pos}{\mathbf{e}}
\newcommand{\objects}{\mathbf{O}}
\newcommand{\centroids}{\mathbf{C}}
\newcommand{\cost}{\mathbf{T}_{\text{c}}}
\newcommand{\loc}{\boldsymbol{z}_{t,k}}
\newcommand{\p}{\boldsymbol{p}}
\newcommand{\PP}{\mathbf{P}}
\newcommand{\Q}{\mathbf{Q}}
\newcommand{\A}{\mathbf{A}}
\newcommand{\Y}{\mathbf{Y}}
\newcommand{\mask}{\mathbf{M}}
\newcommand{\tmarg}{\mathbf{r}}
\newcommand{\cmarg}{\mathbf{c}}
\newcommand{\cls}{\texttt{[CLS]}}
\newcommand{\model}{g}
\newcommand{\backbone}{f}
\newcommand{\head}{h}
\newcommand{\overbar}[1]{\mkern 1.5mu\overline{\mkern-1.5mu#1\mkern-1.5mu}\mkern 1.5mu}

\newcommand{\Raa}[1]{\textcolor{red}{#1}}
\newcommand{\Ra}{\Raa{V8gM} }
\newcommand{\Rbb}[1]{\textcolor{green}{#1}}
\newcommand{\Rb}{\Rbb{HvqY} }
\newcommand{\Rcc}[1]{\textcolor{blue}{#1}}
\newcommand{\Rc}{\Rcc{B8sr} }
\newcommand{\Rdd}[1]{\textcolor{orange}{#1}}
\newcommand{\Rd}{\Rdd{R4} }

\title{TeSLA: Test-Time Self-Learning With Automatic Adversarial Augmentation}

\author{
{\centering Devavrat Tomar$^{1}$ \quad Guillaume Vray$^{1}$ \quad Behzad Bozorgtabar$^{1,2}$ \quad Jean-Philippe Thiran$^{1,2}$}\\
$^{1}$EPFL \quad $^{2}$CHUV\\
{\tt\small $^{1}$\{firstname\}.\{lastname\}@epfl.ch}
}

\thispagestyle{plain}
\pagestyle{plain}

\maketitle

\begin{abstract}
Most recent test-time adaptation methods focus on only classification tasks, use specialized network architectures, destroy model calibration or rely on lightweight information from the source domain. To tackle these issues, this paper proposes a novel \textbf{Te}st-time \textbf{S}elf-\textbf{L}earning method with automatic \textbf{A}dversarial augmentation dubbed \textbf{TeSLA} for adapting a pre-trained source model to the unlabeled streaming test data. In contrast to conventional self-learning methods based on cross-entropy, we introduce a new test-time loss function through an implicitly tight connection with the mutual information and online knowledge distillation. Furthermore, we propose a learnable efficient adversarial augmentation module that further enhances online knowledge distillation by simulating high entropy augmented images. Our method achieves state-of-the-art classification and segmentation results on several benchmarks and types of domain shifts, particularly on challenging measurement shifts of medical images. TeSLA also benefits from several desirable properties compared to competing methods in terms of calibration, uncertainty metrics, insensitivity to model architectures, and source training strategies, all supported by extensive ablations. Our code and models are available on \href{https://github.com/devavratTomar/TeSLA}{GitHub}.\vspace{-0.5em}
\end{abstract}

\section{Introduction}\label{sec:intro}
Deep neural networks (DNNs) perform exceptionally well when the training (\textit{source}) and test (\textit{target}) data follow the same distribution. However, distribution shifts are inevitable in real-world settings and propose a major challenge to the performance of deep networks after deployment. Also, access to the labeled training data may be infeasible at test time due to privacy concerns or transmission bandwidth. In such scenarios, \textbf{source-free domain adaptation} (\textbf{SFDA}) \cite{li2020model,agarwal2022unsupervised,kundu2020universal} and \textbf{test-time adaptation} (\textbf{TTA}) methods \cite{su2022revisiting,iwasawa2021test,liu2021ttt} aim to adapt the pre-trained source model to the unlabeled distributionally shifted target domain while easing access to source data. While SFDA methods have access to all full target data through multiple training epochs (offline setup), TTA methods usually process test images in an online streaming fashion and represent a more realistic domain adaptation. However, most of these methods are applied: (i) only to classification tasks, (ii) evaluated on the non-real-world domain shifts, e.g., the non-measurement shift; (iii) destroy model calibration—entropy minimizing with overconfident predictions \cite{wang2021tent} on incorrectly classified samples, and (iv) use specialized network architectures or rely on the source dataset feature statistics \cite{liu2021ttt}. 

\begin{figure}[t]
    \centering
    \includegraphics[width=\linewidth]{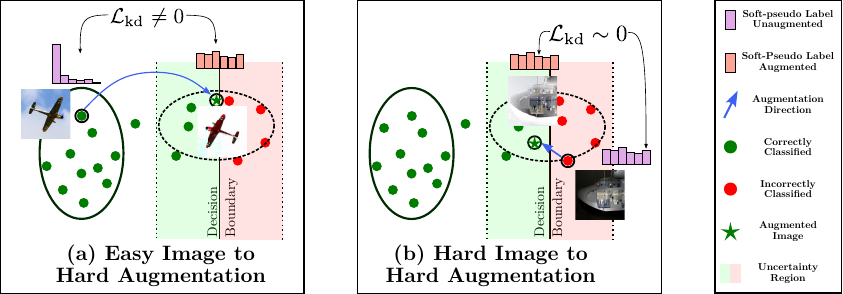}
    \caption{\textbf{Knowledge distillation with adversarial augmentations.} \textbf{(a)} \textit{Easy} images with confident soft-pseudo labels and \textbf{(b)} \textit{Hard} images with unconfident soft-pseudo labels are adversarially augmented and pushed to the uncertainty region (high entropy) near the decision boundary. The model is updated for \textbf{(a)} to match its output on the augmented views with non-augmented views of \textit{Easy} test images using KL-Divergence $\mathcal{L}_\text{kd}\neq0$, while not updated for \textbf{(b)} as $\mathcal{L}_\text{kd}\sim0$ between \textit{Hard} images and their augmented views.
    \vspace{-0.5em}}
    \label{fig:kl_adver_aug}
\end{figure}


We address these issues by proposing a new test-time adaptation method with automatic adversarial augmentation called \textbf{TeSLA}, under which we further define realistic TTA protocols. Self-learning methods often supervise the model adaptation on the unlabeled test images using their predicted pseudo-labels. As the model can easily overfit on its own pseudo-labels, a weight-averaged teacher
model (slowly updated by the student model) is employed for obtaining the pseudo-labels \cite{tarvainen2017mean,wang2022continual}. The student model is then trained with cross-entropy ($\mathbb{CE}$) loss between the one-hot pseudo-labels and its predictions on the test images. In this paper, we instead propose to minimize flipped cross-entropy between the student model's predictions and the soft pseudo-labels (notice the reverse order) with the negative entropy of its marginalized predictions over the test images. In Sec.~\ref{sec:method}, we show that the proposed formulation is an equivalence to mutual information maximization implicitly corrected by the teacher-student knowledge distillation via pseudo-labels, yielding performance improvement on various test-time adaptation protocols and compared to the basic $\mathbb{CE}$ optimization (cf. ablation in Fig. \ref{fig:ablation_components}-(1)).


Motivated by teacher-student knowledge distillation, another central tenet of our method is to assist the student model during adaptation in improving its performance on the hard-to-classify (high entropy) test images. For this purpose, we propose learning \textbf{automatic adversarial augmentations} (see Fig. \ref{fig:kl_adver_aug}) as a proxy for simulating images in the uncertainty region of the feature space. The model is then updated to ensure consistency between {predictions} of high entropy augmented images and soft-pseudo labels from the respective non-augmented versions. Consequently, the model is self-distilled on \textit{Easy} test images with confident soft-pseudo labels (Fig. \ref{fig:kl_adver_aug}\textcolor{red}{a}). In contrast, the model update on \textit{Hard} test images is discarded (Fig. \ref{fig:kl_adver_aug}\textcolor{red}{b}), resulting in better class-wise feature separation. 
In summary, our contributions are: (i) we propose a novel test-time self-learning method based on \textit{flipped} cross-entropy (\textit{f-$\mathbb{CE}$}) through the tight connection with the mutual information between the model’s predictions and the test images; (ii) we propose an efficient \textit{plug-in} test-time automatic adversarial augmentation module used for online knowledge distillation from the teacher to the student network that consistently improves the performance of test-time adaptation methods, including ours; and (iii) TeSLA achieves new state-of-the-art results on several benchmarks, from common image corruption to realistic measurement shifts for classification and segmentation tasks. Furthermore, TeSLA outperforms existing TTA methods in terms of calibration and uncertainty metrics while making no assumptions about the network architecture and source domain's information, e.g., feature statistics or training strategy.

\section{Related Work}
\label{sec:rel_work}
In general, domain adaptation methods aim to distill knowledge from source data that are well-generalizable to target data and relax the assumption of i.i.d. between source and target datasets. To circumvent expensive and cumbersome annotation of new target data, \textbf{unsupervised domain adaptation (UDA)} emerges, and there has been a large corpus of UDAs\cite{peng2019moment,zhang2020collaborative,hoffman2018cycada,ganin2015unsupervised} to match the distribution of domains on both labeled source data and unlabeled target data.  Nonetheless, the above-mentioned methods demand source data to achieve the domain adaptation process, which is often impractical in real-world scenarios, e.g., due to privacy restrictions. The above issue motivates research into \textbf{source-free domain adaptation (SFDA)} \cite{liang2020we,li2020model,agarwal2022unsupervised,kundu2020universal} and \textbf{test-time training} or \textbf{adaptation (TTA)} \cite{su2022revisiting,iwasawa2021test,liu2021ttt,wang2021tent}, which are more closely relevant to our problem setup.

SFDA approaches formulate domain adaptation through pseudo labeling \cite{liang2020we,kundu2020universal}, target feature clustering \cite{yang2021generalized}, synthesizing extra training samples \cite{kundu2020towards}, or feature restoration \cite{eastwood2022sourcefree}. SHOT \cite{liang2020we} proposed feature clustering via information maximization while incorporating pseudo-labeling as additional supervision. BAIT \cite{yang2020casting} leverages the fixed source classifier as source anchors and uses them to achieve feature alignment between the source and target domain. CPGA \cite{Qiu2021CPGA} proposed the SFDA method based on matching feature prototypes between source and target domains. AaD-SFDA \cite{yang2022local} proposed to optimize an objective by encouraging prediction consistency of local neighbors in feature space. Nevertheless, SFDA methods require apriori access to all target data in advance, and the current pseudo-labeling approach, e.g., SHOT \cite{liang2020we}, used offline pseudo-label refinement on a per-epoch basis. In a more realistic domain adaptation scenario, SFDA will still be incompetent to perform inference and adaptation simultaneously.  

TTA methods \cite{wang2021tent,chen2022contrastive} propose alleviating the domina shift by online  (or streaming) adapting a model at test time. Still, we argue that there are ambiguities over the problem setup of TTA in the literature, particularly on whether sequential inference on target test data is feasible upon arrival \cite{iwasawa2021test,wang2021tent} or whether training objectives must be adapted \cite{liu2021ttt,sun2020test}. TTT \cite{sun2020test} proposed fine-tuning the model parameters via rotation classification task as a proxy. On-target adaptation \cite{wang2021target} used pseudo-labeling and contrastive learning to initialize the target-domain feature; each performed independently in their method. T3A \cite{iwasawa2021test} utilized pseudo-labeling to adjust the classifier prototype. More recently, TTAC \cite{su2022revisiting} leveraged the clustering scheme to match target domain clusters to source domain categories. Nevertheless, existing TTA methods rely on specialized neural network architectures or only update a fraction of model weights yielding limited performance gain on the target data. For instance, TENT \cite{wang2021tent} proposed updating affine parameters in the batchnorm layers of convolutional neural networks (CNN), while AdaContrast \cite{chen2022contrastive} and SHOT \cite{liang2020we} used an additional weight normalization classification layer with a projection head. In contrast, we show our methods’ superiority for various neural network architectures, including CNNs and vision transformers (ViTs) \cite{dosovitskiy2020vit}, without additional architectural requirements and different source model training strategies. Moreover, we update all model parameters, and our method is stable over a wide range of hyper-parameters, e.g., learning rate.

\textbf{Test time augmentation} methods \cite{lyzhov2020greedy,ashukha2019pitfalls} are another popular line of domain adaptation research. 
GPS \cite{lyzhov2020greedy} learns optimal augmentation sub-policies by combining image transformations of RandAugment \cite{cubuk2020randaugment} that minimize calibrated log-likelihood loss \cite{ashukha2019pitfalls} on the validation set. OptTTA \cite{tomar2022opttta} optimized the magnitudes of augmentation sub-policies using gradient descent to maximize mutual information and match feature statistics over augmented images. Though effective, these methods \cite{lyzhov2020greedy, tomar2022opttta, wang2019aleatoric} are computationally expensive as all augmentation sub-policies need to be evaluated, making their real-time application difficult. Instead, our proposed adversarial augmentation policies can be learned online and are several orders faster than the existing learnable test-time augmentation strategies, e.g., \cite{tomar2022opttta}.





\section{Methodology}\label{sec:method}
\begin{figure*}[t!]
    \centering
    \includegraphics{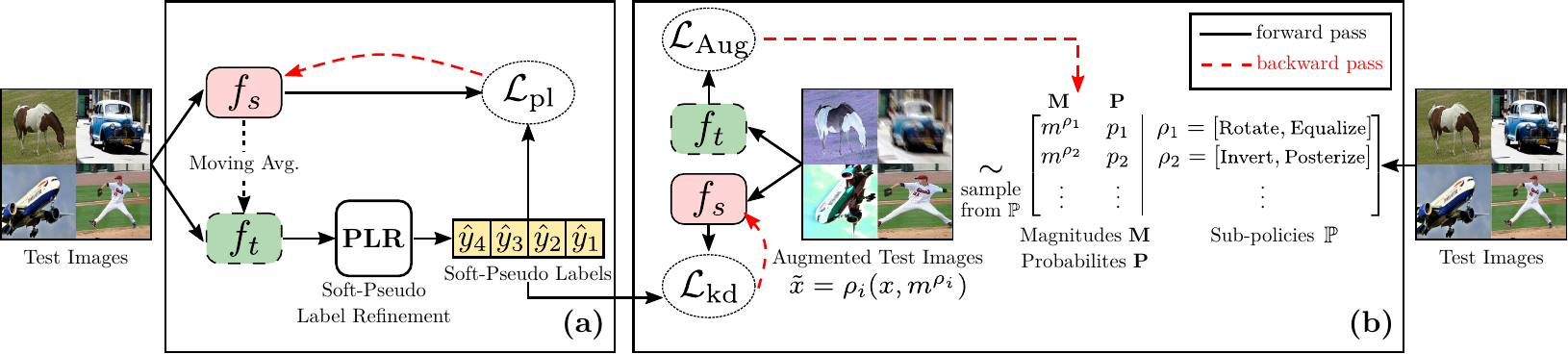}
    \caption{\textbf{Overview of TeSLA Framework.} (a) The student model $f_s$  is adapted on the test images by minimizing the proposed test-time objective $\mathcal{L}_\text{pl}$. The high-quality soft-pseudo labels required by $\mathcal{L}_\text{pl}$ are obtained from the exponentially weighted averaged teacher model $f_t$ and refined using the proposed Soft-Pseudo Label Refinement (PLR) on the corresponding test images. The soft-pseudo labels are further utilized for teacher-student knowledge distillation via $\mathcal{L}_\text{kd}$ on the adversarially augmented views of the test images. (b) The adversarial augmentations are obtained by applying learned sub-policies sampled i.i.d from $\mathbb{P}$ using the probability distribution $P$ with their corresponding magnitudes selected from $M$. The parameters $M$ and $P$ of the augmentation module are updated by the \textit{unbiased gradient estimator} (Eq. \ref{eq:policy_update}) of the loss $\mathcal{L}_\text{Aug}$ computed on the augmented test images. \vspace{-0.5em}}
    \label{fig:ours_main}
\end{figure*}

\noindent\textbf{Overview.} We first introduce our flipped cross-entropy (\textit{f-$\mathbb{CE}$}) loss through the tight connection with the mutual information between the model's predictions and the test images.
(Sec. \ref{sec:sl_mi}). Using this equivalence, we derive our test-time loss function that implicitly incorporates teacher-student knowledge distillation. In Sec. \ref{sec:kd_da}, we propose to enhance the test-time teacher-student knowledge distillation by utilizing the consistency of the student model's predictions on the proposed adversarial augmentations with their corresponding refined soft-pseudo labels from the teacher model. We refine the soft-pseudo labels by averaging the teacher model's predictions on (1) weakly augmented test images; (2) nearest neighbors in the feature space. Furthermore, we propose an efficient online algorithm for learning adversarial augmentations (Sec. \ref{sec:leanable_aug}). Finally, in Sec. \ref{sec:TeSLA}, we summarize our online test-time adaptation method, \textbf{TeSLA}, based on self-learning with the proposed automatic adversarial augmentation. The overall framework is shown in Fig. \ref{fig:ours_main}. 

\vspace{0.5em}
\noindent\textbf{Problem setup.} Given an existing model $f_{\theta _{0}}$, parametrized by $\theta _{0}$ and pre-trained on the inaccessible source data, we aim to improve its performance by updating all its parameters using the unlabeled streaming test data $\mathbf{X}$ during adaptation. Motivated by the concept of \textit{mean teacher} \cite{tarvainen2017mean} as the weight-averaged model over training steps that can yield a more accurate model, we build our online knowledge distillation framework upon teacher-student models. We utilize identical network architecture for teacher and student models. Each model $f=h\circ g$ comprises a backbone encoder $g: \mathbf{X}\to \mathbb{R}^D$ mapping unlabeled test image $\im \in \mathbf{X}$ to the feature representation $\dense=g\left (\im  \right )\in \mathbb{R}^D$ and a classifier (hypothesis) head $h: \mathbb{R}^D\to \mathbb{R}^K$ mapping $\dense$ to the class prediction $\ps\in \mathbb{R}^K$, where $D$ and $K$ denote the feature dimension and number of classes. The parameters of teacher $\theta_t$ and student $\theta_s$ are first initialized from the source model parameters $\theta _{0}$. Then the teacher model's parameters $\theta_t$ are updated from the student model's parameters $\theta_s$ using an exponential moving average (EMA) with momentum  coefficient $\alpha$ as: $\theta_t \leftarrow \alpha \cdot \theta_t + (1-\alpha) \cdot \theta_s$.
%
Following \cite{liang2020we}, we freeze the classifier $h$'s parameters and only update the encoder $g$'s parameters during test-time adaptation.
%
\subsection{Rationale Behind the \textit{f}-CE Objective}\label{sec:sl_mi}
We start by analyzing the rationale behind the proposed \textit{f-$\mathbb{CE}$} loss and its benefit for self-learning through the tight connection with the mutual information between the model's predictions and unlabeled test images. Before that, we first define the notations. 

\vspace{0.5em}
\noindent\textbf{Notations.} We define the random variables of unlabeled test images and the predictions from the student model $f_s$ as $\mathbf{X}$ and $\mathbf{Y}$ and those of the teacher model $f_t$’s soft pseudo-label predictions as $\hat{\mathbf{Y}}$. Furthermore, let $p_{\mathbf{Y}}$ denotes the marginal distribution over $\mathbf{Y}$, $p_{\left ( \mathbf{Y},\mathbf{X} \right )}$ be the joint distribution of $\mathbf{Y}$ and $\mathbf{X}$, and $p_{\mathbf{Y}\mid \mathbf{X}}$ be the conditional distribution of $\mathbf{Y}$ given $\mathbf{X}$. Then, the entropy of $\mathbf{Y}$ and the conditional entropy of $\mathbf{Y}$ given $\mathbf{X}$ can be defined as $\mathcal{H}\left ( \mathbf{Y} \right ):= \mathbb{E}_{p_{\mathbf{Y}}}\left [ -\log p_{\mathbf{Y}}\left ( \mathbf{Y} \right )\right ]$ and $\mathcal{H}\left ( \mathbf{Y}\mid \mathbf{X} \right ):= \mathbb{E}_{p_{\left ( \mathbf{Y},\mathbf{X} \right )}}\left [ -\log p_{\mathbf{Y}\mid \mathbf{X} }\left ( \mathbf{Y}\mid \mathbf{X} \right )\right ]$, respectively.
Besides, let the flipped cross-entropy (\textit{f-$\mathbb{CE}$}) given
 $\mathbf{X}$ be $\mathcal{H}\left ( \mathbf{Y}; \hat{\mathbf{Y}}\mid \mathbf{X} \right ):= \mathbb{E}_{{p_{\left ( \mathbf{Y},\mathbf{X} \right )}}}\left [ -\log p_{\hat{\mathbf{Y}}\mid \mathbf{X}}\left ( \mathbf{Y}\mid \mathbf{X} \right )\right ]$.

\vspace{0.5em}
\noindent\textbf{Self-learning and mutual information.} Based on the above definitions, the \textit{f-$\mathbb{CE}$} between $\mathbf{Y}$ and $\hat{\mathbf{Y}}$ conditioned over $\mathbf{X}$ has the tight connection with the mutual information $\mathcal{I}\left ( \mathbf{Y};\mathbf{X} \right )$ between unlabeled test images $\mathbf{X}$ and the predictions from the student model $\mathbf{Y}$ as follows:

\begin{align}
\label{eq1}
\mathcal{H}\left ( \mathbf{Y}; \hat{\mathbf{Y}}\mid \mathbf{X} \right )=\mathcal{H}\left ( \mathbf{Y}\mid \mathbf{X} \right )+\mathcal{D}_{\text{KL}}\left (\mathbf{Y}\parallel\hat{\mathbf{Y}}\mid \mathbf{X}   \right )\\
=-\mathcal{I}\left ( \mathbf{Y};\mathbf{X} \right )+\mathcal{H}\left ( \mathbf{Y} \right )+\mathcal{D}_{\text{KL}}\left (\mathbf{Y}\parallel\hat{\mathbf{Y}}\mid \mathbf{X}   \right ) \nonumber
\end{align}

This implies that minimizing the \textit{f-$\mathbb{CE}$} ($ \mathcal{H}\left ( \mathbf{Y}; \hat{\mathbf{Y}}\mid \mathbf{X} \right )$ in Eq. \ref{eq1}) and maximizing the entropy of class-marginal prediction $\mathcal{H}\left ( \mathbf{Y} \right )$ is equivalent to maximizing the mutual information between the test images and the student model's predictions $\mathcal{I}\left ( \mathbf{Y};\mathbf{X} \right )$ with a correction KL-divergence term $\mathcal{D}_{\text{KL}}\left (\mathbf{Y}\parallel\hat{\mathbf{Y}}\mid \mathbf{X}   \right )$ involving soft-pseudo labels as:

\begin{equation}\label{eq:sl_mu}
    \underbrace{\mathcal{H}\left ( \mathbf{Y}; \hat{\mathbf{Y}}\mid \mathbf{X} \right )}_{\text{\textit{f-$\mathbb{CE}$}}}-\mathcal{H}\left ( \mathbf{Y} \right ) = -\underbrace{\mathcal{I}\left ( \mathbf{Y};\mathbf{X} \right )}_{\text{Mutual Info.}}  +\mathcal{D}_{\text{KL}}\left (\mathbf{Y}\parallel\hat{\mathbf{Y}}\mid \mathbf{X}   \right )
\end{equation} 
Thus, minimizing the left side of Eq. \ref{eq:sl_mu} with respect to the student model's parameters $\theta_s$ allows the student model to cluster test images using mutual information and criterion-corrected knowledge distillation from soft-pseudo labels. Using the above formulation, we define the following test-time objective to train the student model $f_s$ on a batch of $B$ test images $X=\{\im_1, \dots, \im_B\}$ with their corresponding soft-pseudo labels from teacher model $\Hat{Y}=\{\hat{\ps}_1, \dots, \hat{\ps}_B\}$.
\begin{multline}\label{eq:loss_pl}
    \mathcal{L}_\text{pl}(X, \Hat{Y}) = -\frac{1}{B}\sum_{i=1}^{B}\sum_{k=1}^{K}f_s(\im_i)_k\log((\Hat{\ps}_i)_k)\\
    + \sum_{k=1}^{K}\Hat{f_s}(X)_k\log(\Hat{f_s}(X))_k
\end{multline} 
where $\Hat{f_s}(X)=\frac{1}{B}\sum_{i=1}^{B}f_s(\im_i)$ denotes the marginal class distribution over the batch of test images $X$.


\subsection{Knowledge Distillation via Augmentation}\label{sec:kd_da}
\noindent\textbf{Self-learning from adversarial augmentations.} As illustrated in Fig. \ref{fig:kl_adver_aug}, adversarial augmentations are learned by pushing their feature representations toward the decision boundary (\textit{Expansion} phase), followed by updating the student model $f_s$ to match its prediction on these augmented images with their corresponding soft-pseudo label (\textit{Separation} phase), yielding better separation of features into their respective classes. In our method setup, we continually learn automatic augmentations of the test images (Sec. \ref{sec:leanable_aug}) that are \textbf{adversarial} to the current teacher model $f_t$ and enforce consistency between the student model’s $f_s$ predictions on the augmented views and the corresponding soft-pseudo labels. Since we freeze the classifier module, $f_s$ and $f_t$ share the common decision boundary. Let $\Tilde{\im}$ denote the learned adversarial augmented view of $\im$ and $\Hat{\ps}$ denote the corresponding soft-pseudo label. We minimize the following knowledge distillation loss $\mathcal{L}_\text{kd}$ using KL-divergence to distill knowledge from the teacher model to the student model:
\begin{equation}\label{eq:loss_kd}
    \mathcal{L}_\text{kd}(\Tilde{\im}, \Hat{\ps}) = \mathcal{D}_\text{KL}(\Hat{\ps}\| f_s(\Tilde{\im})) 
\end{equation}

\vspace{0.5em}
\noindent\textbf{Soft pseudo label refinement (PLR).} We refine the quality of soft pseudo-labels and the feature representations by averaging the teacher model's outputs on multiple weakly augmented image views $\rho_w(\im)$ (via {\footnotesize\textsc{{flipping, cropping}}} augmentations) of the same test image $\im$ as follows:

\begin{equation}\label{eq:weak_aug_ensemble}
        \dense_t, \ps_t \leftarrow \mathbb{E}_{\denseu \in \rho_w(\im)}[g_t(\denseu), h_t(g_t(\denseu))]
\end{equation} 
The refined \textit{feature representations} $\dense_t$ and the \textit{softmaxed pseudo labels} $\ps_t$ from the encoder $g_t$ and the classifier $h_t$ are further stored in an online memory queue $\mathbf{Q}$ of fixed size. The final soft pseudo-label is computed by averaging the refined soft pseudo-labels of $n$-nearest neighbors of the current test image in the feature space as follows:

\begin{equation}\label{eq:neighbors}
\begin{gathered}
    \mathbf{Q}[\arg\max (\ps_t)] \text{.append}(\{\dense_t, \ps_t\}) \\
    \Hat{\ps} = \frac{1}{n}\sum \mathcal{N}_{\mathbf{Q},n}(\dense_t)
\end{gathered}
\end{equation}
where $\mathbf{Q}$ denotes an online memory of class-balanced queues (with size $|\mathbf{Q}[\arg\max (\ps_t)]|\leq N_\mathbf{Q}$) of the refined feature representations and the softmaxed label predictions on the previously seen test images, and $\mathcal{N}_{\mathbf{Q},n}(\dense_t)$ denotes soft labels of $n$-nearest neighbors of $\dense_t$ from $\mathbf{Q}$.

\subsection{Learning Adversarial Data Augmentation}\label{sec:leanable_aug}
This section presents our efficient online method for learning adversarial data augmentation for the current teacher $f_{t}$. We first introduce our adversarial augmentation search space. Then, we describe our differentiable strategy to optimize and sample adversarial augmentations that push the feature representations of the test images toward the uncertain region in the close vicinity of the decision boundary.

\vspace{0.5em}
\noindent\textbf{Policy search space.} Let $\mathbb{O}$ be a set of all image transformation operations $\mathcal{O}:\mathbb{R}^{H \times W \times 3} \rightarrow \mathbb{R}^{H \times W \times 3}$ defined in our search space. In particular, we use image transformations {\footnotesize\textsc{{Auto-Contrast, Equalize, Invert, Solarize, Posterize, Contrast, Brightness, Color, ShearX, ShearY, TranslateX, TranslateY, Rotate, and Sharpness}}}. Each transformation $\mathcal{O}$ is specified with its magnitude parameter $m\in[0,1]$.
Since the output of some image operations may not be conditional on its magnitude (e.g., {\footnotesize\textsc{{Equalize}}}) or may not be differentiable
(e.g., {\footnotesize\textsc{{Posterize}}}), for such image operations, we use the straight-through gradient estimate w.r.t. each pixel as: $\partial \mathcal{O}\left ( \im_{ij} \right )/\partial m = \mathbf{1}$ for magnitude optimization.

We define a sub-policy $\rho$ as a combination of $N$ image operations (sub-policy dimension) from $\mathbb{O}$ that is applied sequentially to a given image $\im$ as:
\begin{equation}
\rho\left ( \im, m^{\rho}\right )=\mathcal{O}_{1}^{\rho}\cdots \mathcal{O}_{N}^{\rho}\left ( \im,m_{N}^{\rho} \right )
\end{equation}
where $m^\rho=[m_1^\rho, \dots, m_N^\rho]$ denotes the magnitude set of the sub-policy $\rho$. We also define $\mathbb{P}=\{\rho_1, \rho_2, \dots\}$ as the set of all possible sub-policies.

\vspace{0.5em}
\noindent\textbf{Policy evaluation.} We propose learnable adversarial augmentation, aiming to optimize and sample data augmentations as a proxy for simulating data in the uncertain region of the feature space. Given the teacher model $f_t$, a sub-policy $\rho$ with magnitude $m$ is evaluated for a test image $\im$ using the following adversarial objective:

\begin{equation}\label{eq:policy_eval}
    \mathcal{L}_\text{aug} (\boldsymbol{x}, \rho) = \sum_{k=1}^K f_{t}\left (\tilde{\boldsymbol{x}}  \right )\log \left (f_{t}\left (\tilde{\boldsymbol{x}}  \right )  \right )+ \lambda_1 r(\tilde{\boldsymbol{x}} , \boldsymbol{x}) 
\end{equation}
where $\tilde{\boldsymbol{x}}=\rho\left ( \boldsymbol{x},m \right )$ denotes the adversarial augmented image and $r\left ( \cdot,\cdot   \right )$ regularizes augmentation severity. The hyperparameter $\lambda_1$ controls the augmentation severity. We optimize the adversarial augmentation loss $\mathcal{L}_\text{aug}$ with respect to the sub-policy $\rho$'s parameters. Minimizing $\mathcal{L}_\text{aug}$ is equivalent to maximizing the entropy of the teacher model's prediction on the augmented image $\Tilde{\im}$, thus pushing its feature representation towards the decision boundary (first term). For the augmentation regularization (second term), we use the mean squared distance function between the teacher encoder's $L$ internal layers' activations of the adversarial augmentation and non-augmented versions of image $\im$ defined as:

\begin{equation}
    r(\tilde{\boldsymbol{x}}, \im) = \frac{1}{L}\sum_{l=1}^L\|\mu_l(\tilde{\boldsymbol{x}}) - \mu_l(\im)\|^2
\end{equation} 
where $\mu_{l}\left ( \cdot  \right )$ denotes the mean activation of the $l^{th}$ layer of the teacher encoder $g_t$.

\vspace{0.5em}
\noindent\textbf{Policy optimization.} Let $M=[m^{\rho_1},\dots,m^{\rho_{\|\mathbb{P}\|}}]$ denote the set of magnitudes $m^\rho$ of all sub-policies $\rho\in\mathbb{P}$ and $P=[p_1, \dots, p_{\|\mathbb{P}\|}]$ denote the probability of selecting a sub-policy from $\mathbb{P}$. The expected policy evaluation loss (Eq. \ref{eq:policy_eval}) for an image $\im$ over the policy search space is given by:

\begin{equation}
    \mathbb{E}[\mathcal{L}_\text{aug} (\im)] = \sum_{i=1}^{\|\mathbb{P}\|} p_i\cdot\mathcal{L}_\text{aug} (\im, \rho_i)
\end{equation}
Evaluating the gradient of $\mathbb{E}[\mathcal{L}_\text{aug}(\im)]$ w.r.t. $M$ and $P$ can become computationally expensive as $\|\mathbb{P}\|\sim \binom{\|\mathbb{O}\|}{N}$. Thus, we use the following re-parameterization trick to estimate its \textit{unbiased gradient}:

\begin{align}
\label{eq:unbiased}
\nabla\mathbb{E}[\mathcal{L}_\text{aug}(x)] = \delta(\im, \mathbb{P}) = \sum_{i=1}^{\|\mathbb{P}\|}\nabla(p_i\cdot\mathcal{L}_\text{aug}(\im, \rho_i))\\
=\sum_{i=1}^{\|\mathbb{P}\|}p_i(\nabla\mathcal{L}_\text{aug}(\im, \rho_i) + \mathcal{L}_\text{aug}(\im, \rho_i)\cdot\nabla \log p_i)\nonumber
\end{align} 
Thus, an unbiased estimator of $\delta(\im, \mathbb{P})$ can be written as:
\begin{equation}
    \Hat{\delta}(\im, \rho_i) = \nabla\mathcal{L}_\text{aug}(\im, \rho_i) + \mathcal{L}_\text{aug}(\im, \rho_i)\cdot\nabla \log p_i
\end{equation} 
where index $i$ is sampled from the probability distribution $P$. For an online batch of $B$ test images $\{\im_1, \im_2, \dots, \im_B\}$, we apply augmentation sub-policies $\{\rho_{i_1}, \rho_{i_2} \dots, \rho_{i_B}\}$ from $\mathbb{P}$ where $\{i_1, i_2, \dots, i_B\}$ are sampled i.i.d from distribution $P$, and update the parameters $P$ and $M$ with the following stochastic gradient update rule:
\begin{equation}\label{eq:policy_update}
    [P, M] \leftarrow [P, M] - \frac{\gamma}{B}\sum_{j=1}^B\Hat{\delta}(\im_j, \rho_{i_j})
\end{equation} 
where $\gamma$ is the learning rate. We set $\gamma=0.1$ for all experiments without the need for hyperparameter tuning. 


\subsection{Self-Learning With Adversarial Augmentation}\label{sec:TeSLA}
We minimize the following overall objective $\mathcal{L}_\text{TeSLA}$ for training the student model $f_s$ on a batch of $B$ test images $X=\{\im_1, \dots, \im_B\}$ and their adversarial augmented views $\Tilde{X}=\{\Tilde{\im}_1, \dots, \Tilde{\im}_B\}$ with the corresponding refined soft-pseudo labels from the teacher model $\Hat{Y}=\{\hat{\ps}_1, \dots, \hat{\ps}_B\}$:

\begin{equation}\label{eq:loss_TeSLA}
    \mathcal{L}_\text{TeSLA}(X,\Tilde{X}, \Hat{Y}) = \mathcal{L}_\text{pl}(X,\Hat{Y}) + \frac{\lambda_2}{B}\sum_{i=1}^{B} \mathcal{L}_\text{kd}(\Tilde{\im}_i, \Hat{\ps}_i)
\end{equation} 
where $\lambda_2$ is a hyper-parameter for the knowledge distillation. The adversarial augmented views $\Tilde{X}$ are obtained by sampling \textit{i.i.d.} augmentation sub-policies from $\mathbb{P}$ for every test image in $X$ with probability $P$ and applying the corresponding magnitude from $M$. Concurrently, we also optimize $M$ and $P$ using the policy optimization mentioned in Sec. \ref{sec:leanable_aug}.

\section{Experiments}\label{sec:exps}

\subsection{Datasets and Experimental Settings}\label{sec:datasets}
We evaluate and compare TeSLA against state-of-the-art (SOTA) test-time adaptation algorithms for both classification and segmentation tasks under three types of test-time distribution shifts resulting from (1) \textbf{common image corruption}, (2) \textbf{synthetic to real data transfer}, and (3) \textbf{measurement shifts} on medical images. The latter is characterized by a change in medical imaging systems, e.g., different scanners across hospitals or various staining techniques.  

\noindent\textbf{TTA protocols.} We adopt the TTA protocols in \cite{su2022revisiting} and categorize competing methods based on two factors: (i) source training objective and (ii) sequential or non-sequential inference. First, unlike \cite{su2022revisiting}, we use \textbf{Y} to indicate if access to the source domain's information, e.g., feature statistics is possible or if the source training objective is allowed to be modified; otherwise, we use \textbf{N}. Next, we use \textbf{O} to indicate a one-pass adaptation and evaluation protocol (one epoch) for sequential test data and \textbf{M} to show a multi-pass adaptation on all test data and inference after several epochs.
Thus, we end up with four possible TTA protocols, namely \textbf{N-M}, \textbf{Y-M}, \textbf{N-O}, and \textbf{Y-O}. Unlike previous works \cite{liu2021ttt,su2022revisiting}, TeSLA does not rely on any source domain feature distribution, yielding two realistic TTA protocols: \textbf{N-M} and \textbf{N-O}.

\noindent\textbf{Hyperparameters for all experiments.} We set $\lambda _{1}=\lambda _{2}=1.0$ for all experiments (cf. ablation in \textbf{Appendix~\apxE}). For the adversarial augmentation module, we use sub-policy dimension $N=2$ (except for VisDA-C \cite{visda2017}, $N=4/3$ for N-O/N-M protocols) (cf. \textbf{Appendix~\apxE}) and Adam optimizer \cite{kingma2014adam}. For pseudo-label refinement (PLR), we use five weak augmentations composed of random flips and resize crop.
More details of hyperparameters used for each experiment can be found in \textbf{Appendix~\apxA}.

\noindent\textbf{Common image corruptions.}
To evaluate TeSLA's efficacy for the classification task, we use \textbf{CIFAR10-C/CIFAR100-C} \cite{hendrycks2018benchmarking} and large-scale \textbf{ImageNet-C} \cite{hendrycks2018benchmarking} datasets each containing 19 types of corruptions applied to the clean test set with five levels of severity. We perform validation on 4 out of 19 types of corruption {\footnotesize\textsc{{(Spatter, Gaussian Blur, Speckle Noise, Saturate)}}} to select hyperparameters and test on the remaining 15 corruptions at the maximum severity level of 5. Following \cite{liu2021ttt}, we train the ResNet50 \cite{he2016deep} on the clean CIFAR10/CIFAR100/ImageNet training set and adapt it to classify the unlabeled corrupted test set.
\noindent\textbf{Synthetic to real data adaption.} We use challenging and large-scale \textbf{VisDA-C}, and \textbf{VisDA-S} \cite{visda2017} datasets for evaluating synthetic-to-real data adaptation at test-time for classification and segmentation tasks, respectively. Following \cite{chen2022contrastive, liang2020we}, we adapt the ResNet101 network pre-trained on synthetic images to classify 12 vehicle classes on the photo-realistic images of VisDA-C. While for VisDA-S, we adapt the DeepLab-v3 \cite{chen2017rethinking} backbone pre-trained on synthetic GTA5 \cite{Richter_2016_ECCV} to the Cityscapes \cite{Cordts2016Cityscapes} to segment 19 classes.

\noindent\textbf{Measurement shifts on medical images.} We access TeSLA on staining variations for the tissue type classification of hematoxylin \&  eosin (H\&E) stained patches from colorectal cancer tissue slides. We use MobileNetV2 \cite{sandler2018mobilenetv2} trained on the source \textbf{Kather-19} dataset \cite{kather2019predicting} and adapt it to the target \textbf{Kather-16} dataset \cite{kather_2016_53169} on four tissue categories: tumor, stroma, lymphocyte, and mucosa. We also evaluate TeSLA for variations in scanners and imaging protocols in multi-site medical images on two magnetic resonance imaging (MRI) datasets for the segmentation task, namely the \textbf{multi-site prostate MRI} \cite{liu2020ms} and \textbf{spinal cord grey matter segmentation (SCGM)} \cite{prados2017spinal}. Following \cite{tomar2022opttta}, we adapt the U-Net \cite{ronneberger2015u} from site 1 to sites \{2,3,4\} of the spinal cord dataset, and  from sites \{A,B\} to sites \{D,E,F\} of the prostate dataset.



\noindent\textbf{Competing baselines.} We compare TeSLA with the following SFDA and TTA baselines, including direct inference of the trained source model on the target test data without adaptation (\textbf{Source}). We also implement the SFDA-based pseudo-labeling baselines from a basic pseudo-labeling approach (\textbf{PL}) to the \textbf{SHOT} approach~\cite{liang2020we} based on training the feature extraction module by adopting the information maximization loss to make globally diverse but individually certain predictions on the target domain. For recent TTA methods, we compare our method against \textbf{TTT++}~\cite{liu2021ttt}, \textbf{AdaContrast}~\cite{chen2022contrastive}, \textbf{CoTTA}~\cite{wang2022continual}, basic \textbf{TENT}~\cite{wang2021tent} and \textbf{BN}~\cite{ioffe2015batch} as representative methods based on feature distribution alignment via stored statistics, contrastive learning, augmentation-averaged predictions, entropy minimization, and batch normalization statistic. Furthermore, we compare TeSLA to more recent methods based on test-time augmentation policy learning, \textbf{OptTTA} \cite{tomar2022opttta}, and anchor clustering \textbf{TTAC}~\cite{su2022revisiting}. By default, ~\cite{liang2020we,ioffe2015batch,chen2022contrastive} follow the N-M protocol, and we adapt them to the N-O protocol, while ~\cite{wang2021tent,tomar2022opttta,wang2022continual}, by default, follow the N-O protocol and adapt them to the N-M protocol setting. Using source features distribution, TTAC and TTT++ follow Y-O and Y-M.
We report error rates on classification tasks and Mean Intersection over Union (mIoU)/ Dice scores on segmentation tasks.
For consistency across baselines and TTA protocols, we do not use any specialized model architectures (e.g., projection head, weight-normalized classifier layer) for SHOT~\cite{liang2020we} and AdaContrast~\cite{chen2022contrastive}.

\begin{figure*}[t]
      \centering
      \begin{subfigure}{0.4\textwidth}
          \centering
          \includegraphics[width=\linewidth]{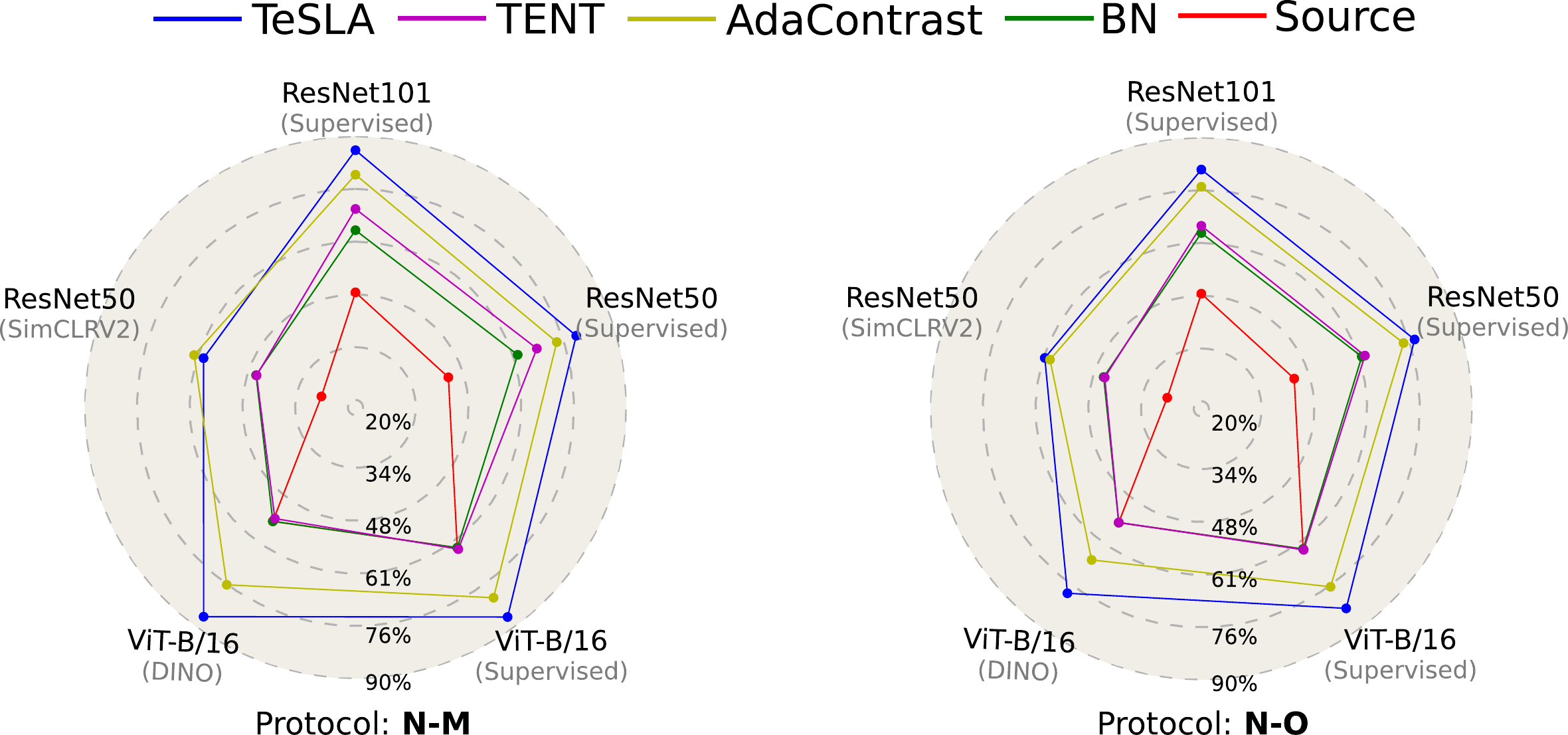}
          \caption{}
          \label{fig:radar_plot}
      \end{subfigure}
      \hspace{1em}
      \begin{subfigure}{0.4\textwidth}
          \centering
          \includegraphics[width=\linewidth]{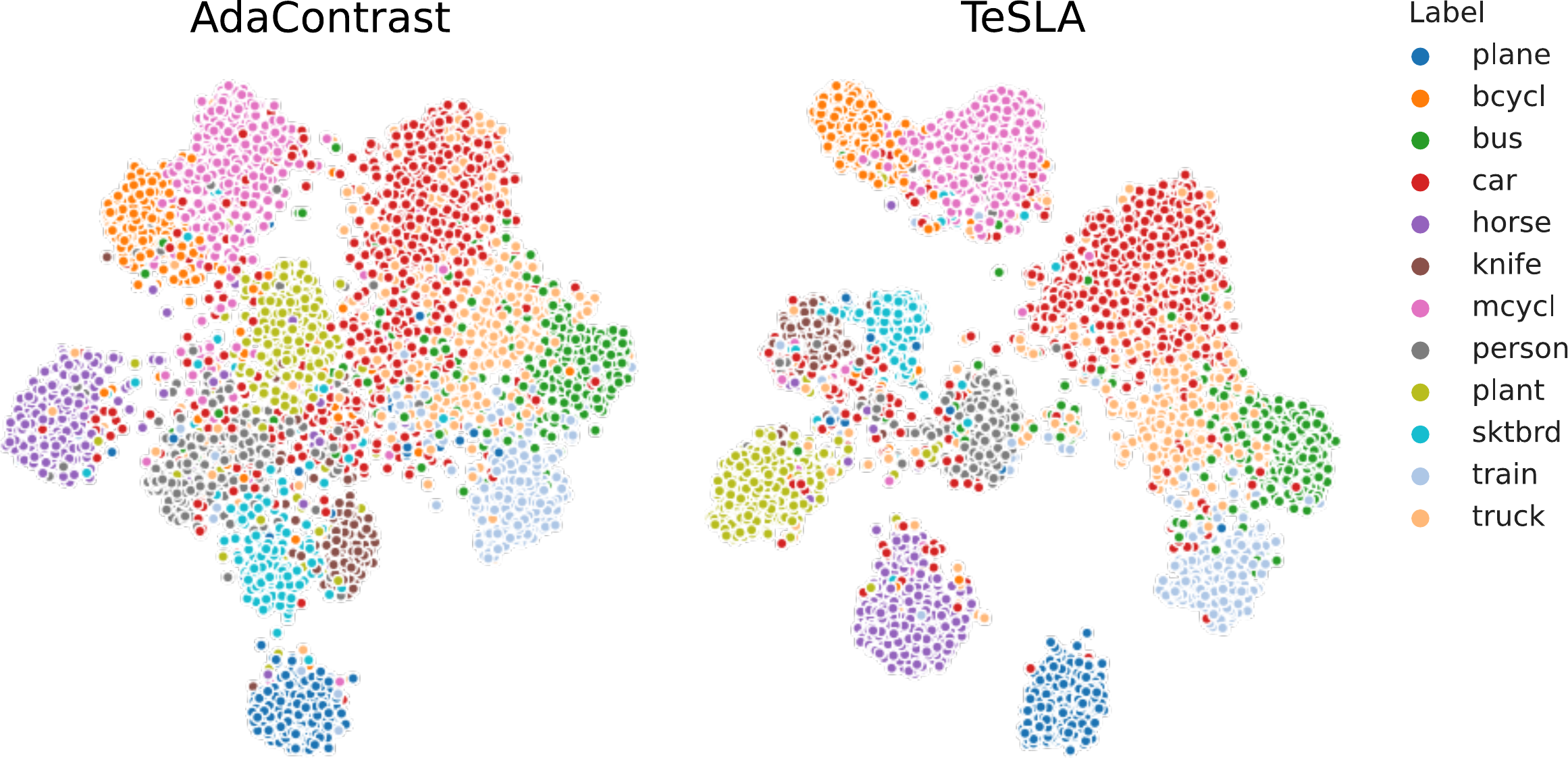}
          \caption{}
          \label{fig:tsne_plot}
      \end{subfigure}
    \label{fig:results_callibration_tsne}
    \caption{\textbf{Ablation experiments on the VisDA-C}. \textbf{(a) Source training strategy and model architecture.} TeSLA outperforms the competing TTA methods across varied source training strategies (Supervised, SimCLRV2\cite{chen2020big}, and DINO\cite{caron2021emerging}) and model architectures (ResNet50, ResNet101 \cite{he2016deep}, and ViT-B/16\cite{dosovitskiy2020vit}) using the same set of hyperparameters for both protocols (N-O/N-M). Each vertex represents the mean class avg accuracy over 3 seeds. \textbf{(b) t-SNE visualization} \cite{van2008visualizing} comparison of feature embeddings for AdaContrast \cite{chen2022contrastive} and TeSLA.}
    \vspace{-0.5em}
\end{figure*}

\begin{table}[t]
    \centering
    \caption{\textbf{Comparison of SOTA TTA methods under different protocols} evaluated on CIFAR-10/100-C, ImageNet-C, VisDA-C and Kather-16 datasets. We report the average error computed over 15 test corruptions for the common image corruption shifts. We also report \textit{Class Avg.} in \% error rates for synthetic-to-real and measurement shifts over 3 and 10 seeds, respectively.}
    
    \resizebox{\linewidth}{!}{
    \begin{tabular}{@{}c|c|cccccc|cc|cc@{}}
    
    \toprule
    
       \multirow{3}[3]{*}{Method} &  \multirow{3}[3]{*}{\rotatebox[origin=c]{90}{Protocol}} & \multicolumn{6}{c|}{\cellcolor[HTML]{FFFFE0}\textbf{Common Image Corruptions}} & \multicolumn{2}{c|}{\cellcolor[HTML]{FFE0FF}\textbf{Syn-to-Real}} & \multicolumn{2}{c}{\cellcolor[HTML]{FED8B1}\makecell{\textbf{Measurement}\\\textbf{Shift}}} \\
       
      \cmidrule(lr){3-8} \cmidrule(lr){9-10} \cmidrule(lr){11-12}
       
      \multicolumn{1}{c|}{} & \multicolumn{1}{c}{} & \multicolumn{2}{|c}{CIFAR10-C} & \multicolumn{2}{c}{CIFAR100-C} & 
      \multicolumn{2}{c|}{ImageNet-C} &
      \multicolumn{2}{c|}{VisDA-C} & \multicolumn{2}{c}{Kather-16} \\
      
      \cmidrule(lr){3-4} \cmidrule(lr){5-6} \cmidrule(lr){7-8} \cmidrule(lr){9-10} \cmidrule(lr){11-12}
      
       & & \multicolumn{1}{c}{O} & M & O & M & O & \multicolumn{1}{c|}{M} & O & M & O & M \\ 

    \midrule

    Source & N & \multicolumn{2}{c}{29.1} & \multicolumn{2}{c}{60.4} & \multicolumn{2}{c|}{81.8} & \multicolumn{2}{c|}{51.5} & \multicolumn{2}{c}{32.0} \\
    
    \midrule
    
    BN~\cite{nado2020evaluating,ioffe2015batch}& N & 15.6 & 15.4 & 43.7 & 43.3 & 67.7 & 67.6 & 35.4 & 35.0 & 18.3 & 18.2\\
    TENT~\cite{wang2021tent}& N & 14.1 & 12.9 & 39.0 & 36.5 & 57.4 & 54.2 & 33.5 & 29.3 & 16.2 & 12.0 \\
    SHOT~\cite{liang2020we} & N & 13.9 & 14.2 & 39.2 & 38.7 & 68.7 & 68.2 & 29.4 & 24.5 & 14.7 & 12.0 \\
    AdaContrast~\cite{chen2022contrastive} & N & \multicolumn{6}{c|}{-} & 23.1 & 20.2 & \multicolumn{2}{c}{-}\\
    
    \midrule
    
    TTT++ \cite{liu2021ttt} & Y & 15.8 & 9.8 & 44.4 & 34.1 & 59.3 & - & 35.2 & 34.1 & 16.7 & 7.9\\
    TTAC~\cite{su2022revisiting} & Y & 13.4 & \textbf{9.4} & 41.7 & 33.6 & 58.7 & - &  32.2 & 31.1 & 9.6 & 5.5\\ 
    
    \midrule
    
    \rowcolor[HTML]{E0FFFF}
     \textbf{TeSLA} & N & 12.5 & 9.7 & 38.2 & 32.9 & 55.0 & \textbf{51.5} & \textbf{17.8} & \textbf{13.5} & \textbf{9.2} & 3.3 \\
     
     \rowcolor[HTML]{E0FFFF}
     \textbf{TeSLA-s} & Y & \textbf{12.1} & 9.7 & \textbf{37.3} & \textbf{32.6}
     & \textbf{53.1} & - & 24.0 & 17.9 & 9.9 & \textbf{3.1} \\

    \bottomrule
    
    \end{tabular}}
    \label{tab:results_classification}
    \vspace{-0.5em}
\end{table}

\subsection{Results}\label{subsec:results}
\noindent \textbf{Classification task.} In Table \ref{tab:results_classification}, we summarize the average (Avg) classification error rates (\%) of TeSLA against several SOTA \textbf{SFDA} and \textbf{TTA} baselines on the \textbf{common image corruptions} of CIFAR10-C, CIFAR100-C, ImageNet-C; \textbf{synthetic to real data} domain shifts of VisDA-C; and \textbf{measurement shifts} on the Kather-16 dataset. We also report per-class top-1 accuracies for VisDA-C and Kather-16 datasets and corruption-wise error rates on CIFAR-10-C/CIFAR-100-C and ImageNet-C datasets in \textbf{Appendix~\apxB}. TeSLA easily surpasses all competing methods under the protocols N-O and N-M on all the datasets. In particular, TeSLA outperforms the second-best baseline under N-O/N-M protocol by a (\%) margin 1.4/3.2 on CIFAR10-C, 0.8/3.6 on CIFAR100-C, 2.4/2.7 on ImageNet-C, 5.3/6.7 on VisDA-C, and 5.5/8.7 on the Kather-16 datasets. Despite not utilizing any source feature statistic, TeSLA (N-O/N-M) is either competitive or surpasses TTAC\cite{su2022revisiting} and TTT++\cite{liu2021ttt} (Y-O/Y-M) that benefit from the source dataset feature statistics during adaptation.
Including the global source feature alignment module of TTAC with TeSLA (\textbf{TeSLA-s}) could further improve its performance under (Y-O/Y-M) protocols for slight domain shifts (e.g., common image corruptions). We report the results of AdaContrast for VisDA-C only as the implementation for other datasets are not provided. We do not report TeSLA-s results on ImageNet-C under the Y-M protocol, as previous baselines were evaluated only Y-O protocol.


\begin{figure}[t]
  \centering
  \includegraphics[width=\linewidth]{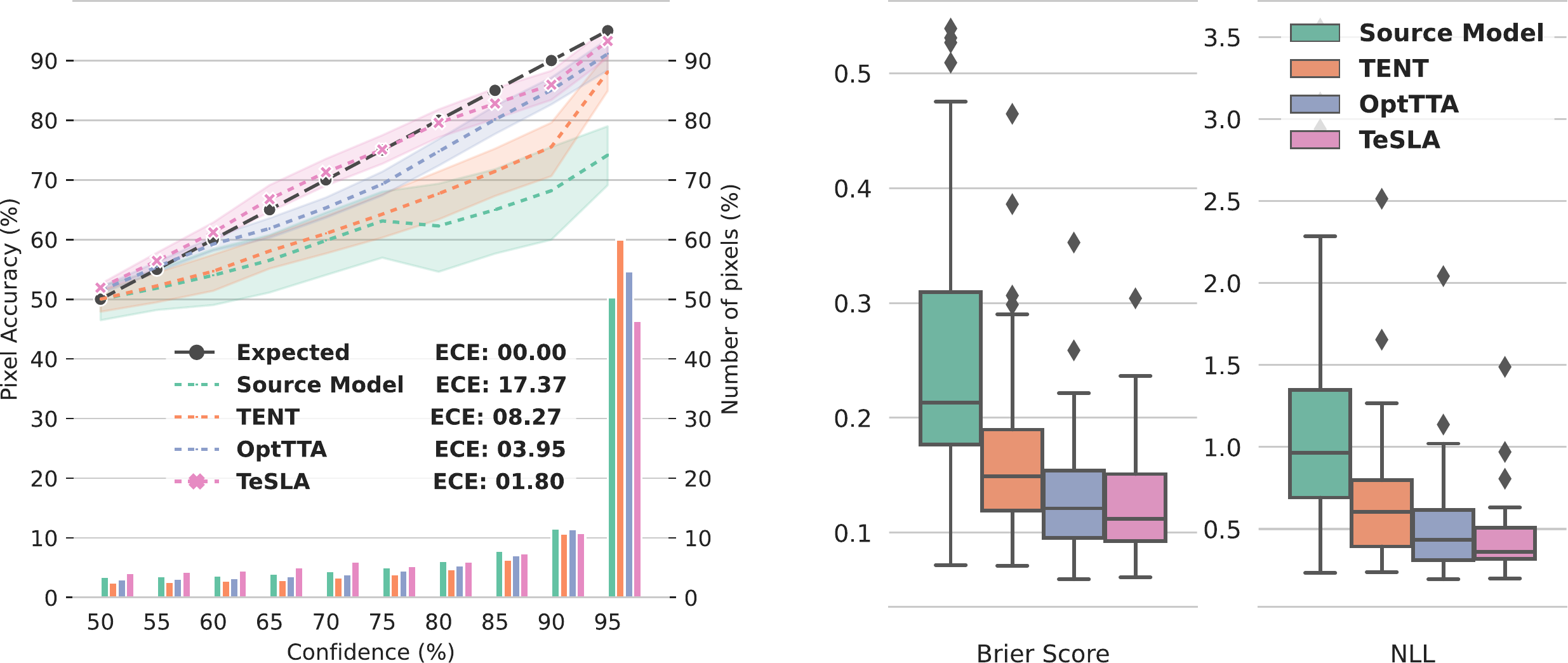}
  \caption{\textbf{Model calibration and uncertainty estimation.} Model calibration using reliability diagrams \cite{niculescu2005predicting} and model uncertainty estimation using Brier Score and Negative Log Likelihood (NLL) of the Source Model, TENT\cite{wang2021tent}, OptTTA\cite{tomar2022opttta} and, TeSLA on the prostate dataset for test-time adaptation. \vspace{-0.5em}}
  \label{fig:reliability_brier_plot}
\end{figure}

\begin{table}[t]
    \centering
    \caption{\textbf{Semantic segmentation results} (\textit{Class Avg.} mIoU in \%) on the VisDA-S (GTA5 $\rightarrow $ Cityscapes) test-time adaptation task. }
    \vspace{-0.5em}
    \resizebox{0.9\linewidth}{!}{
    \begin{tabular}{c|c|c|ccccc}
    
    \toprule
    
    \multicolumn{2}{c|}{Protocol} & Source & BN\cite{nado2020evaluating} & TENT\cite{wang2021tent} & PL & CoTTA\cite{wang2022continual} & \cellcolor[HTML]{E0FFFF}\textbf{TeSLA} \\
    
    \midrule
    
    N & O & \multirow{2}[2]{*}{35.3}& 36.7 & 38.3 & 38.8 & 37.0 & \cellcolor[HTML]{E0FFFF}\textbf{44.5} \\
    
    N & M & & 38.4 & 39.2 & 38.6 & 39.9 & \cellcolor[HTML]{E0FFFF}\textbf{46.0} \\
    
    \bottomrule
    
    \end{tabular}}
    \vspace{-0.5em}
    \label{tab:results_gta_cityscape}
\end{table}

\begin{table}[ht]
    \centering
    \caption{\textbf{Semantic segmentation results} (\textit{Class Avg.} volume-wise \textit{mean} Dice  in \%) on the cross-site spinal cord and prostate MRI test-time adaptation tasks. }
    \resizebox{\linewidth}{!}{
    \begin{tabular}{c|c|c|ccccc}
    
    \toprule
    
    \multicolumn{2}{c|}{Protocol} & Source & BN\cite{nado2020evaluating} & TENT\cite{wang2021tent} & PL & OptTTA\cite{tomar2022opttta} & \cellcolor[HTML]{E0FFFF}\textbf{TeSLA} \\
    
    \midrule
    \multicolumn{8}{l}{\textbf{Spinal Cord}\hspace{1em}$\texttt{Site}\left \{ \texttt{1} \right \}\rightarrow \texttt{Sites}\left \{\texttt{2,3,4}  \right \}$ } \\
    \midrule
    
    N & O & \multirow{2}[2]{*}{76.0\std{11.8}} & 81.6\std{8.3} & 81.1\std{9.1} & 81.7\std{8.6} & 84.1\std{4.8} & \cellcolor[HTML]{E0FFFF}\textbf{85.3\std{5.8}} \\
    
    N & M & & 84.3\std{4.8} & 84.4\std{4.7} & 84.3\std{4.7} & 84.3\std{4.4} & \cellcolor[HTML]{E0FFFF}\textbf{85.4\std{4.4}} \\
    
    \midrule
    \multicolumn{8}{l}{\textbf{Prostate}\hspace{1em}$\texttt{Sites}\left \{ \texttt{A,B} \right \}\rightarrow \texttt{Sites}\left \{\texttt{D,E,F}  \right \}$} \\
    \midrule

    N & O & \multirow{2}[2]{*}{60.5\std{27.0}} & 72.1\std{15.2} & 74.7\std{17.9} & 72.4\std{15.2} & 83.1\std{7.7} & \cellcolor[HTML]{E0FFFF}\textbf{83.5\std{6.5}} \\ 
    N & M & & 73.1\std{18.0} & 81.2\std{9.3} & 81.1\std{9.2} & 83.4\std{7.7} & \cellcolor[HTML]{E0FFFF}\textbf{84.3\std{5.8}} \\
    
    \bottomrule
    
    \end{tabular}}
    \label{tab:results_mri}
    \vspace{-0.5em}
\end{table}

\noindent \textbf{Segmentation task.} Unlike TeSLA, several TTA methods e.g., AdaContrast\cite{chen2022contrastive}, SHOT\cite{liang2020we}, TTT++\cite{liu2021ttt}, TTAC\cite{su2022revisiting}, have \textit{only} been evaluated on the classification task. Therefore, we compare TeSLA against the current SOTA \textbf{TTA} methods applied to the segmentation task on \textbf{synthetic-to-real data transfer} of the VisDA-S dataset in Table \ref{tab:results_gta_cityscape}. TeSLA significantly outperforms the competing methods and achieves the best mIOU scores for both N-O and N-M protocols, beating the second-best baseline by a margin of +5.7\% and +6.1\%, respectively. In Table \ref{tab:results_mri}, we compare TeSLA against the recent SOTA test-time augmentation policy method, OptTTA\cite{tomar2022opttta} and other TTA baselines for the inter-site adaptation on two challenging MRI datasets mentioned in Sec. \ref{sec:datasets}. TeSLA convincingly outperforms all other methods on severe measurement shifts across sites under $\textbf{N-O}$ and $\textbf{N-M}$ protocols.

\noindent \textbf{Computational cost for adversarial augmentations.} As shown in Table \ref{tab:computation_cost}, TeSLA learns the adversarial augmentation in an online manner, and thus its runtime is several orders faster than learnable test-time augmentation OptTTA \cite{tomar2022opttta}, while comparable to that of static augmentation policies with an additional overhead of only 0.10 GPU hours/epoch on the VisDA-C dataset.


\noindent \textbf{Test-time feature visualization.} Fig.~\ref{fig:tsne_plot} compares t-SNE projection \cite{van2008visualizing} of the encoder's features of AdaContrast\cite{chen2022contrastive} with TeSLA on the VisDA-C. TeSLA shows better inter-class separation for features than AdaContrast, as supported by an improved silhouette score from $0.149$ to $0.271$.

\noindent \textbf{Model calibration and uncertainty estimation.} Entropy minimization-based TTA methods \cite{wang2021tent, liang2020we} explicitly make the model confident in their predictions, resulting in poor model calibration. For trusting the adapted model, it should output reliable confidence estimates matching its true underlying performance on the test images. Such a well-behaved model is characterized by expected \textbf{calibration error (ECE)} through \textbf{reliability diagram} \cite{niculescu2005predicting}, and uncertainty metrics of \textbf{brier score} and \textbf{negative log-likelihood (NLL)} \cite{gomariz2021probabilistic}. In Fig. \ref{fig:reliability_brier_plot}, we plot the reliability diagram (dividing the probability range $\left [ 0,0.5 \right ]$ into 10 bins), and uncertainty metrics of the Source model, TENT\cite{wang2021tent} and OptTTA\cite{tomar2022opttta} against TeSLA for the inter-site model adaptation on the prostate dataset \cite{liu2020ms}. We observe that TeSLA achieves the best model calibration (lowest ECE of 1.80\%) and lowest Brier and NLL scores of 0.12 and 0.24 ($p$ $<$ 0.006 against TENT).




\begin{table}[t]
    \centering
    \caption{\textbf{Performance comparison of adversarial augmentation} learned by TeSLA during adaptation against the prior art augmentation methods on runtime (GPU hours/epoch on {NVIDIA GeForce RTX 3090}) and \textit{Class Avg.} accuracy in \% on the VisDA-C dataset.}
    \vspace{-0.5em}
    \resizebox{\linewidth}{!}{
    \begin{tabular}{c|cccc|c}
    \toprule
         Augmentation Type & OptTTA & \makecell{TeSLA\\RA\cite{cubuk2020randaugment}} & \makecell{TeSLA\\AA\cite{cubuk2018autoaugment}} & \cellcolor[HTML]{E0FFFF}\makecell{TeSLA\\$N=2$} & \cellcolor[HTML]{E0FFFF}\makecell{TeSLA\\$N=3$} \\
         \midrule
         Runtime & 4.50 & 0.05 & 0.06 & \cellcolor[HTML]{E0FFFF}0.16 & \cellcolor[HTML]{E0FFFF}0.18 \\
        \textit{Class Avg.} Acc. (N-O) & - & 80.2 & 81.2 & \cellcolor[HTML]{E0FFFF}81.5 & \cellcolor[HTML]{E0FFFF}82.2 \\
        \textit{Class Avg.} Acc. (N-M) & - & 84.7 & 85.7 & \cellcolor[HTML]{E0FFFF}86.3 & \cellcolor[HTML]{E0FFFF}86.5 \\
     \bottomrule
    \end{tabular}
    }
    \vspace{-0.5em}
    \label{tab:computation_cost}
\end{table}

\subsection{Ablation Studies}\label{sec:ablations}
In this section, we scrutinize the roles played by different components of TeSLA. All ablations are performed on the synthetic-to-real test data adaptation of the VisDA-C dataset.

\begin{figure}[t]
    \centering
    \includegraphics[width=0.32\linewidth]{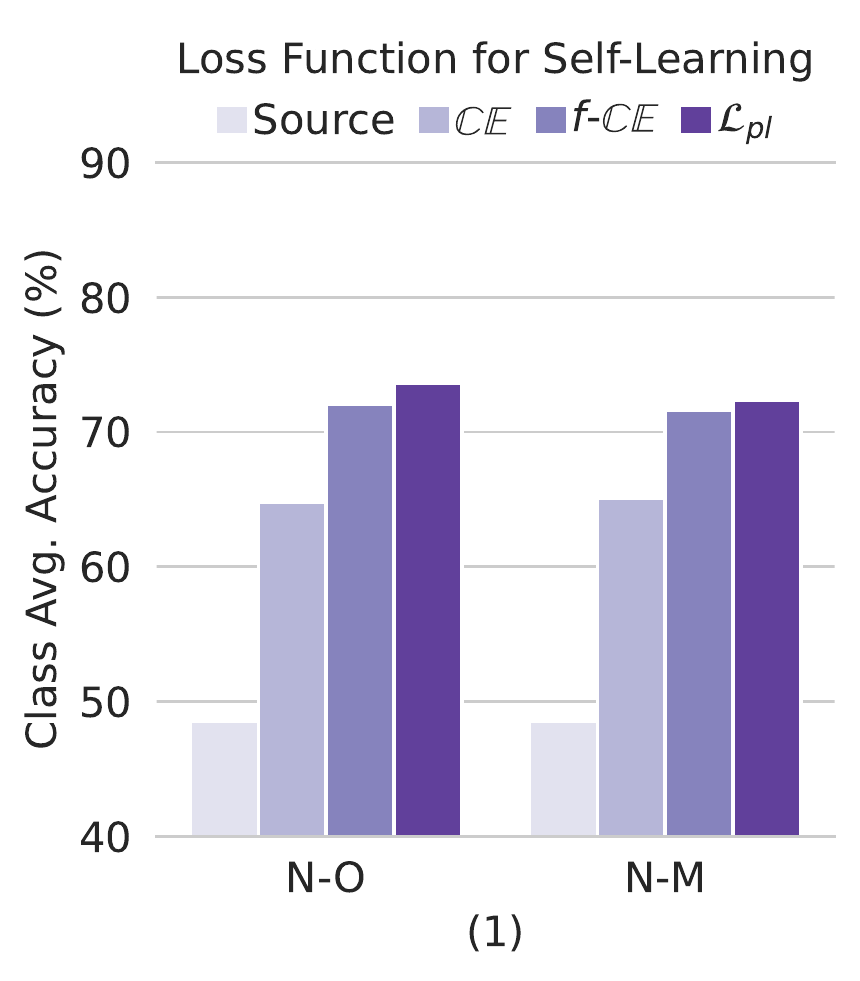}
    \includegraphics[width=0.32\linewidth]{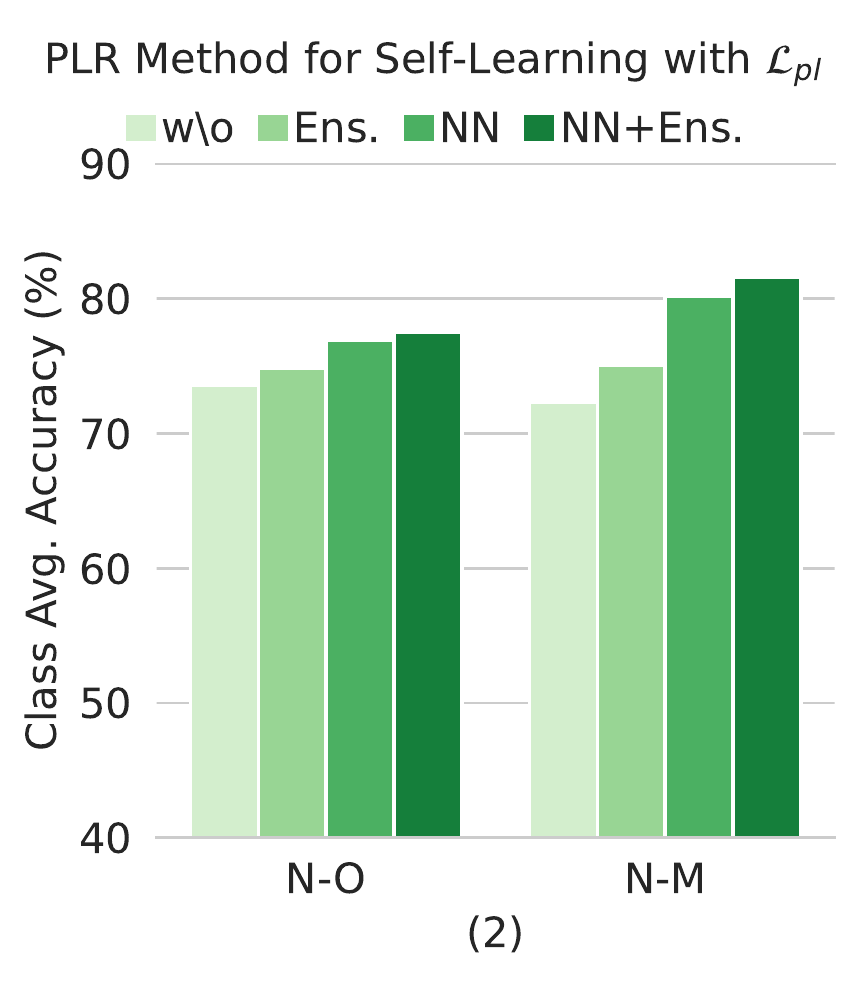}
    \includegraphics[width=0.32\linewidth]{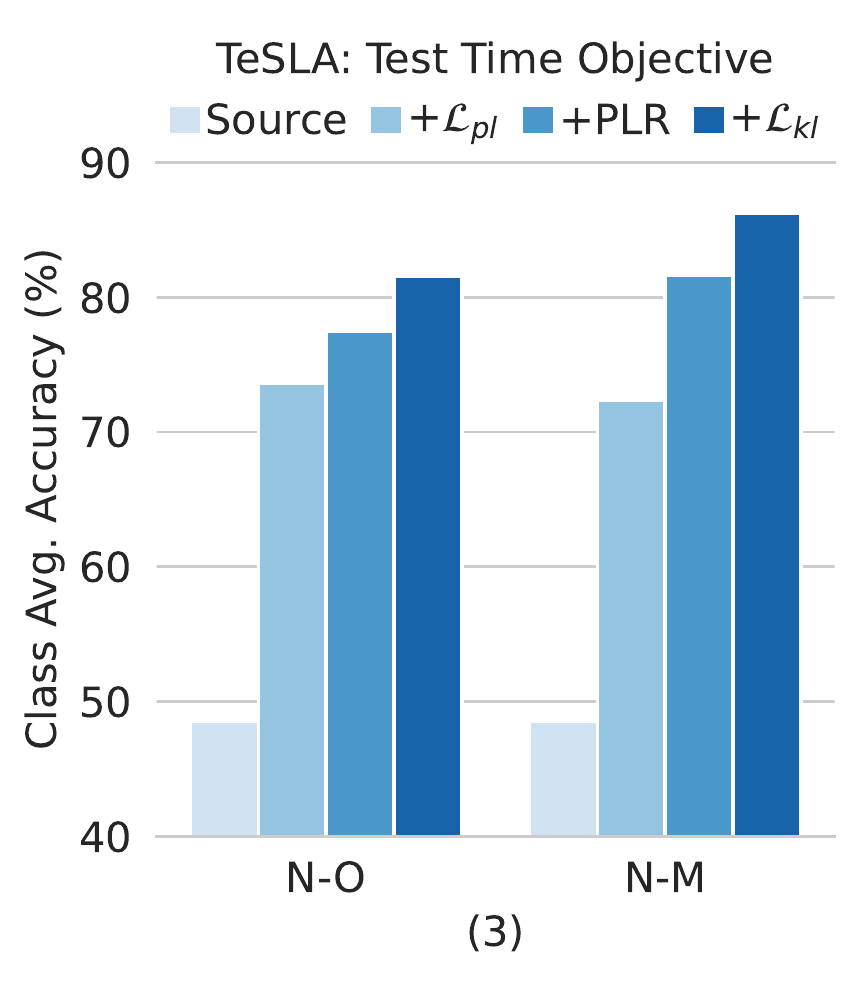}
    \vspace{-0.5em}
    \caption{\textbf{Ablation experiments on the VisDA-C.} (\textit{left to right}) (1) Ablation results for self-learning loss functions, including \textit{f-$\mathbb{CE}$}, $\mathbb{CE}$, and $\mathcal{L}_\text{pl}$; (2) ablation on different PLR methods: ensembling predictions on 5 weak augmentations, averaging predictions of the 10 nearest neighbors in the feature space; (3) ablation on different components of TeSLA's objective function: $\mathcal{L}_\text{pl}$, PLR, and $\mathcal{L}_\text{kd}$.}
    \label{fig:ablation_components}
    \vspace{-0.5em}
\end{figure}

\noindent\textbf{Loss functions for self-learning.} We compare TTA performance of our self-learning objective $\mathcal{L}_\text{pl}$ with the proposed flipped cross-entropy loss \textit{f}-$\mathbb{CE}$ and the basic cross-entropy loss $\mathbb{CE}$ in Fig. \ref{fig:ablation_components}-(1). We do not use the proposed augmentation module and pseudo-label refinement in this experiment. We observe that \textit{f}-$\mathbb{CE}$ alone improves the accuracy of the Source model in the N-M protocol by +23.1\% compared to +16.6\% improvement by $\mathbb{CE}$. Using $\mathcal{L}_\text{pl}$ (Eq. \ref{eq:loss_pl}) further improves the accuracy to a margin of +23.8\%. 

\noindent\textbf{Contribution of individual components: PLR, $\mathcal{L}_\text{pl}$, and $\mathcal{L}_\text{kd}$.} Our test-time objective $\mathcal{L}_\text{TeSLA}$ has three components-- (i) self-learning loss $\mathcal{L}_\text{pl}$, (ii) pseudo-label refinement (PLR), and (iii) knowledge distillation $\mathcal{L}_\text{kd}$ with adversarial augmentation. In Fig. \ref{fig:ablation_components}-(3), we study the effect of accruing individual components to $\mathcal{L}_\text{TeSLA}$  (starting from source trained model) for the test-time adaptation accuracy. We observe that $\mathcal{L}_\text{pl}$ alone improves the model's accuracy from 48.5\% to 72.3\% while adding PLR and $\mathcal{L}_\text{kd}$ further boosts it to 81.6\% (+9.3) and 86.3\% (+4.7), respectively.

\noindent\textbf{PLR: ensembling on weak augmentations (Ens) and nearest neighbor averaging (NN).} We also conduct an ablation on the two pseudo-label refinement approaches -- (i) ensembling the teacher model's output on five weak augmentations. (ii) averaging soft pseudo labels of $n=10$ nearest neighbors in the feature space (cf. sensitivity test in \textbf{Appendix~\apxE}). Fig. \ref{fig:ablation_components}-(2) shows that combining both approaches gives better results than using them individually.

\noindent\textbf{Adversarial augmentation.} Table \ref{tab:computation_cost} shows the benefit of using our adversarial augmentation ($N=2,3$) with TeSLA against (i) \textbf{RandAugment (RA)} \cite{cubuk2020randaugment}, (ii) \textbf{AutoAugment (AA)} \cite{cubuk2019autoaugment} (optimized for ImageNet), and (iii) OptTTA \cite{tomar2022opttta}. We achieve far better \textit{Class Avg.} accuracy on the VisDA-C at the cost of slight runtime overhead compared to static augmentation policies for protocols (N-O, N-M). In addition, TeSLA augmentation (TeAA) consistently improves the performance of other TTA methods (Table \ref{tab:tesla_shot}).
\begin{table}[ht]
    \centering
    \caption{\textit{Class Avg. accuracy (\%)} of \textbf{TeSLA augmentation (TeAA)} with other TTA objectives under N-O protocol on VisDA-C dataset.}
    \vspace{-0.5em}
    \resizebox{0.9\linewidth}{!}{\begin{tabular}{c|c||c|c||c|c}
    \toprule
         TENT & \textbf{+TeAA} & SHOT & \textbf{+TeAA} & AdaContrast & \textbf{+TeAA}\\ 
    \midrule
         66.5 & \textbf{72.0} & 70.6 & \textbf{73.1} & 76.9 & \textbf{79.1} \\
    \bottomrule
    \end{tabular}}
    \vspace{-0.5em}
    \label{tab:tesla_shot}
\end{table}


\noindent\textbf{Source training strategies and model architectures.} The ablation results (Fig. \ref{fig:radar_plot}) on the VisDA-C show that TeSLA surpasses competing TTA baselines for different source training strategies, including \textbf{Supervised} and self-supervised (\textbf{SimCLRV2} \cite{chen2020big}, \textbf{DINO}\cite{caron2021emerging}). Moreover, we ablate TeSLA over different architectures, ranging from ResNet-50/101\cite{he2016deep} to the recent vision transformer ViT-B \cite{dosovitskiy2020vit} using the same set of hyperparameters, showing the merits of no reliance on the network architecture or source training strategy. 


\section{Conclusion}
\label{sec:limitation}
We introduced TeSLA, a novel self-learning algorithm for test-time adaptation that utilizes automatic adversarial augmentation. TeSLA is agnostic to the model architecture and source training strategies and gives better model calibration than other TTA methods. Through extensive experiments on various domain shifts (measurement, common corruption, synthetic to real), we show TeSLA's superiority over previous TTA methods on classification and segmentation tasks. Note that TeSLA assumes class uniformly when implicitly maximizing mutual information and can be improved by incorporating prior, e.g., class label distribution statistics.

 {\small
 \bibliographystyle{ieee_fullname}
 \bibliography{egbib}
 }
 \clearpage
\appendix
\noindent
\textbf{Overview.} This Appendix provides important additional details about our proposed method \textbf{TeSLA}. In Appendix~\ref{appendix:hyperparameters}, we provide hyperparameters details for each test-time adaptation experiment on the \textbf{common image corruptions}, \textbf{synthetic-to-real}, and \textbf{medical measurement shift} datasets. In Appendix~\ref{appendix:additional_results}, we provide additional quantitative results, including class top-1 accuracies for the VisDA-C \cite{visda2017} and Kather-16 \cite{kather_2016_53169} datasets and corruption-wise error rates on the CIFAR-10-C/CIFAR-100-C \cite{hendrycks2018benchmarking}, and ImageNet-C \cite{hendrycks2018benchmarking} datasets. In addition, we also provide segmentation class-wise mean Intersection over Union (mIoU) for the VisDA-S dataset \cite{visda2017} and class average Dice score for different sites of the target test domain on the spinal cord \cite{prados2017spinal} and prostate dataset \cite{liu2020ms} for the competing test time adaptation methods. All quantitative results are included for both one-pass (\textbf{O}) and multi-pass (\textbf{M}) protocols. Appendix~\ref{appendix:runtime} provides an overall runtime computation cost of TeSLA along with other Test Time adaptation methods on VisDA-C \cite{visda2017} dataset, while Appendix~\ref{appendix:other_tt_obj} discusses TeSLA's equivalence to TENT \cite{wang2021tent} and \cite{liang2020we} without mean teacher and adversarial augmentations. We include additional ablation experiments and hyperparameter sensitivity tests in Appendix~\ref{appendix:sensitivity}. Finally, we provide other qualitative results, including a sanity check on TeSLA's adversarial augmentations, uncertainty evaluation, and segmentation visualization in Appendix~\ref{appendix:qualitative_results}. 

\section{Hyperparameter Settings}\label{appendix:hyperparameters}
\renewcommand\thefigure{\thesection.\arabic{figure}}
\renewcommand\thetable{\thesection.\arabic{table}}
\setcounter{figure}{0}
\setcounter{table}{0}
Table \ref{tab:hyperparameters_cls} and Table \ref{tab:hyperparameters_seg}  present the hyperparameters' values of TeSLA used for individual experiments on different classification and segmentation datasets, respectively. These hyperparameters include the batch size $B$, learning rate, optimizer, EMA momentum coefficient $\alpha$, number of epochs for test-time adaptation (M protocol), the number of weak augmentations $|\rho_w|$; the number of nearest neighbors $n$; class-wise queue size $N_Q$ used by soft pseudo-label refinement (PLR) module, number of image operations for augmentation sub-policy $N$ used by the adversarial augmentation module, the augmentation severity controller coefficient $\lambda_1$ and the knowledge distillation coefficient $\lambda_2$ for $\mathcal{L}_\text{kd}$ loss term.

\begin{table}[ht]
\centering
\caption{\textbf{Hyperparameter setting} used for the proposed methods TeSLA/TeSLA-s on different classification datasets.}
\resizebox{\linewidth}{!}{
\begin{tabular}{@{}c|cccccccccc@{}}

    \toprule
     \multirow{2}[2]{*}{\begin{tabular}[c]{@{}c@{}}Hyperparameters\end{tabular}}
     & \multicolumn{2}{c}{CIFAR10-C}
     & \multicolumn{2}{c}{CIFAR100-C}
     & \multicolumn{2}{c}{ImageNet-C}
     & \multicolumn{2}{c}{VisDA-C}
     & \multicolumn{2}{c}{Kather-16}\\
     
     \cmidrule(lr){2-3} \cmidrule(lr){4-5} \cmidrule(lr){6-7} \cmidrule(lr){8-9} \cmidrule(lr){10-11}
     
     & O & M & O & M & O & M & O & M & O & M  \\
     
     \midrule
     
     Batch size $B$ & 128 & 128 & 128 & 128 & 128 & 128 & 128 & 128 & 32 & 32  \\
     
     \midrule
     
     Learning rate & 0.001 & 0.001 & 0.001 & 0.001 & 0.001 & 0.001 & 0.001 & 0.001 & 0.005 & 0.005 \\
     
     \midrule
     
     Optimizer & Adam & Adam & Adam & Adam & SGD & SGD & SGD & SGD & Adam & Adam \\
     
     \midrule
     
    \begin{tabular}[c]{@{}c@{}}Momentum\\coefficient $\alpha$\end{tabular} & 0.99 & 0.999 & 0.99 & 0.999 & 0.9 & 0.996 & 0.9 & 0.996 & 0.9 & 0.96 \\
    
     \midrule
     
    \begin{tabular}[c]{@{}c@{}}Number of\\epochs\end{tabular} & 1 & 70 & 1 & 70 & 1 & 5 & 1 & 5 & 1 & 70 \\
     
     \midrule
     
     \begin{tabular}[c]{@{}c@{}}Number of weak\\augmented views $|\rho_w|$\end{tabular}  & 5 & 5 & 5 & 5 & 5 & 5 & 5 & 5 & 5 & 5 \\
     
     \midrule
     
     \begin{tabular}[c]{@{}c@{}}Number of nearest\\neighbors $n$\end{tabular}  & 1 & 4 & 1 & 1 & 1 & 1 & 10 & 10 & 8 & 8 \\
     
     \midrule
     
    \begin{tabular}[c]{@{}c@{}}Class-wise\\queue size $N_\mathbf{Q}$\end{tabular}  & 1 & 256 & 1 & 1 & 1 & 1 & 256 & 256 & 32 & 256\\
    
    \midrule
    
    \begin{tabular}[c]{@{}c@{}}Sub-policy\\dimension $N$\end{tabular} & 2 & 2 & 2 & 2 & 2 & 2 & 4 & 3 & 2 & 2 \\
    
    \midrule
    
    \begin{tabular}[c]{@{}c@{}}Augmentation severity\\controller $\lambda_1$\end{tabular} & 1 & 1 & 1 & 1 & 1 & 1 & 1 & 1 & 1 & 1 \\
    
    \midrule
    
    \begin{tabular}[c]{@{}c@{}}Knowledge\\distillation weight $\lambda_2$\end{tabular} & 1 & 1 & 1 & 1 & 1 & 1 & 1 & 1 & 1 & 1 \\
    
    \bottomrule
\end{tabular}}
\label{tab:hyperparameters_cls}
\end{table}

\begin{table}[ht]
\centering
\caption{\textbf{Hyperparameter setting} used for the proposed method TeSLA on different segmentation datasets.}
\resizebox{\linewidth}{!}{
\begin{tabular}{@{}c|cccccc@{}}

    \toprule
     \multirow{2}[2]{*}{\begin{tabular}[c]{@{}c@{}}Hyperparameters\end{tabular}}
     & \multicolumn{2}{c}{VisDA-S}
     & \multicolumn{2}{c}{Spinal Cord}
     & \multicolumn{2}{c}{Prostate} \\
     
     \cmidrule(lr){2-3} \cmidrule(lr){4-5} \cmidrule(lr){6-7}
     
     & O & M & O & M & O & M \\
     
     \midrule
     
     Batch size $B$ & 8 & 8 & 16 & 16 & 16 & 16 \\
     
     \midrule
     
     Learning rate  & 0.001 & 0.001  & 0.0002 & 0.0002 & 0.0002 & 0.0002  \\
     
     \midrule
     
      Optimizer & AdamW & AdamW & AdamW & AdamW & AdamW & AdamW \\

     \midrule
     
    \begin{tabular}[c]{@{}c@{}}Momentum\\coefficient $\alpha$\end{tabular} & 0.996 & 0.999 & 0.996 & 0.996 & 0.996 & 0.996\\
    
     \midrule
     
    \begin{tabular}[c]{@{}c@{}}Number of\\epochs\end{tabular} & 1 & 3 & 1 & 5 & 1 & 3\\
     
     \midrule
     
     \begin{tabular}[c]{@{}c@{}}Number of weak\\augmented views $|\rho_w|$\end{tabular} & 3 & 3 & 5 & 5 & 5 & 5 \\
     
     \midrule
     
     \begin{tabular}[c]{@{}c@{}}Number of nearest\\neighbors $n$\end{tabular} & 1 & 1 & 1 & 1 & 1 & 1\\
     
     \midrule
     
    \begin{tabular}[c]{@{}c@{}}Class-wise\\queue size $N_\mathbf{Q}$\end{tabular} & 1 & 1 & 1 & 1 & 1 & 1 \\
    
    \midrule
    
    \begin{tabular}[c]{@{}c@{}}Sub-policy\\dimension $N$\end{tabular} & 3 & 3 & 3 & 3 & 3 & 3\\
    
    \midrule
    
    \begin{tabular}[c]{@{}c@{}}Augmentation severity\\ controller $\lambda_1$\end{tabular} & 1 & 1 & 1 & 1 & 1 & 1 \\
    
    \midrule
    
    \begin{tabular}[c]{@{}c@{}}Knowledge\\distillation weight $\lambda_2$\end{tabular} & 1 & 1 & 1 & 1 & 1 & 1 \\

    \bottomrule
\end{tabular}}
\label{tab:hyperparameters_seg}
\end{table}

\section{Additional Quantitative Results}\label{appendix:additional_results}
\renewcommand\thefigure{\thesection.\arabic{figure}}
\renewcommand\thetable{\thesection.\arabic{table}}
\setcounter{figure}{0}
\setcounter{table}{0}
In Table \ref{tab:kather}, we compare TeSLA against state-of-the-art test-time adaptation methods for the classification task on the Kather-16 dataset. We present the class top-1 accuracies (\%) for each of the four tissue categories of \textbf{tumor}, \textbf{stroma}, \textbf{lymphocyte}, and \textbf{mucosa}. In addition, we report the class average accuracy (Avg.). 

Furthermore, in Table \ref{tab:cifar_c}, we present the corruption-wise average class error rates for different competing test time adaptation baselines, including the proposed TesLA and TeSLA-s on the  CIFAR10-C,
CIFAR100-C and ImageNet-C.  We use the following image corruptions for the evaluation at the maximum severity level of  5: [{\footnotesize\textbf{\textsc{Gaussian Noise, Shot Noise, Impulse Noise, Defocus Blur, Glass Blur, Motion Blur, Zoom Blur, Snow, Frost, Fog, Brightness, Contrast, Elastic Transformation, Pixelate, JPEG Compression}}}]. We also use the ResNet-50 backbone for all experiments.

In Table \ref{tab:visda}, we include the overall and class-wise accuracies for test time adaptation of ResNet-101 trained on synthetic vehicle images (\textbf{training}) and tested on the photo-realistic vehicle images (\textbf{validation}) of the VisDA-C dataset. The photo-realistic images are classified into 12 categories: \textbf{plane}, \textbf{bicycle}, \textbf{bus}, \textbf{car}, \textbf{horse}, \textbf{knife}, \textbf{motor-cycle}, \textbf{person}, \textbf{plant}, \textbf{skate-board}, \textbf{train}, and \textbf{truck}.

In Table \ref{tab:mri}, we present segmentation results (class Avg. volume-wise mean \%Dice score)  for test-time adaptation baselines on two multi-site magnetic resonance imaging (MRI) benchmarks - spinal cord \cite{prados2017spinal} and prostate dataset \cite{liu2020ms}. For the spinal cord dataset, we report results for test-time adaptation of the U-Net segmentation model trained on \textbf{site 1} to \textbf{site 2}, \textbf{site 3}, and \textbf{site 4}. Similarly, we report results of the U-Net segmentation model trained on the sites \textbf{A and B}, which are adapted on the sites \textbf{D}, \textbf{site E}, and \textbf{site F}.

Table \ref{tab:visdas} presents the results of competing test-time adaptation methods applied to the segmentation adaptation task from the synthetic images of GTA \cite{Richter_2016_ECCV} to the photo-realistic images of Cityscapes \cite{Cordts2016Cityscapes} dataset. We report the class-wise mean Intersection over Union (mIoU) over 19 classes: \textbf{road}, \textbf{side-walk}, \textbf{building}, \textbf{wall}, \textbf{fence}, \textbf{pole}, \textbf{light}, \textbf{sign}, \textbf{vegetation}, \textbf{terrain}, \textbf{sky}, \textbf{person}, \textbf{rider}, \textbf{car}, \textbf{truck}, \textbf{bus}, \textbf{train}, \textbf{motor-cycle}, and \textbf{bicycle}.

\begin{table}[t]
\centering
\caption{\textbf{Comparison of state-of-the-art TTA methods under different protocols} on the Kather-16  dataset. We report the class top-1 accuracies (\%) for each of the four classes and the per-class average accuracy (Avg.). Each result is averaged over ten seeds.}
\resizebox{\linewidth}{!}{
\begin{tabular}{c|c|cccc|c}
\toprule
     Method & \rotatebox[origin=c]{90}{Protocol} & tumor & stroma & lymphocyte & mucosa & Avg. \\
     
\midrule

Source & N & 84.5\std{4.0} & 91.6\std{3.0} & 0.9\std{1.2} & 95.0\std{1.3} & 68.0\std{1.3}\\

\midrule

BN & N-O & 89.3\std{2.5} & 85.5\std{2.6} & 61.7\std{2.2} & 90.2\std{0.6} & 81.7\std{1.0}\\

Tent & N-O  & 89.8\std{4.0} & 89.3\std{3.4} & 67.2\std{2.2} & 88.9\std{1.0} & 83.8\std{1.8}\\

SHOT & N-O  & 84.7\std{5.7} & 95.7\std{2.0} & 67.9\std{3.9} & 92.8\std{0.8} & 85.3\std{2.5}\\
 
\midrule

TTT++ & Y-O & 82.8\std{8.5} & 85.1\std{7.6} & 73.7\std{3.8} & 91.4\std{1.7} & 83.3\std{2.7}\\

TTAC & Y-O & 92.6\std{2.6} & 96.6\std{1.9} & \textbf{78.3\std{4.3}} & 93.9\std{1.0} & 90.4\std{1.1}\\

\midrule

TeSLA & N-O  & \textbf{93.5\std{1.8}} & \textbf{98.2\std{1.2}} & 77.3\std{4.1} & \textbf{94.3\std{0.7}} & \textbf{90.8\std{1.1}} \\

TeSLA-s & Y-O & 90.7\std{4.6} & 98.0\std{1.0} & 77.9\std{5.6} & 94.0\std{1.7} & 90.1\std{1.4} \\

\midrule
\midrule

BN & N-M  & 86.3\std{3.6} & 86.1\std{1.9} & 66.9\std{1.2} & 87.7\std{0.7} & 81.8\std{1.0}\\

Tent & N-M & 96.4\std{4.2} & 99.5\std{0.4} & 62.6\std{9.0} & 93.7\std{1.5} & 88.0\std{3.3} \\

SHOT & N-M & 84.6\std{4.4} & 98.5\std{0.6} & 77.1\std{5.2} & 91.7\std{0.8} & 88.0\std{2.4}  \\

\midrule

TTT++ & Y-M & 95.6\std{1.4} & 93.9\std{2.8} & 85.2\std{5.3} & 93.6\std{1.8} & 92.1\std{2.0} \\

TTAC & Y-M & 92.9\std{12.6} & 98.1\std{1.1} & 92.4\std{4.7} & 94.4\std{1.8} & 94.5\std{4.7} \\

\midrule

TeSLA & N-M   & 97.1\std{1.0} & \textbf{99.6\std{0.3}} & 94.4\std{2.0} & 95.6\std{0.9} & 96.7\std{0.5} \\

TeSLA-s & Y-M & \textbf{97.4\std{0.4}} & 99.5\std{0.3} & \textbf{95.1\std{2.0}} & \textbf{95.7\std{1.0}} & \textbf{96.9\std{0.6}} \\

\bottomrule
\end{tabular}}
\label{tab:kather}
\end{table}

\begin{table*}
\centering
\caption{\textbf{Comparison of state-of-the-art TTA methods under different protocols} on common image corruptions datasets, including CIFAR10-C, CIFAR100-C, and ImageNet-C. We report the error rates (\%) on 15 test images' corruptions.}
\resizebox{\linewidth}{!}{
\begin{tabular}{@{}c|c|ccccccccccccccc|c}
\toprule
     Method & \begin{sideways}Protocol\end{sideways} & Gaus. & Shot & Impu. & Defo. & Glas. & Moti. & Zoom & Snow & Fros. & Fog & Brig. & Cont. & Elas. & Pixe. & Jpeg & Avg. \\

\midrule
\multicolumn{18}{c}{\cellcolor[HTML]{EFEFEF}CIFAR10-C} \\
\midrule

Source & N & 48.7 & 44.0 & 57.0 & 11.8 & 50.8 & 23.4 & 10.8 & 21.9 & 28.2 & 29.4 & 7.0 & 13.3 & 23.4 & 47.9 & 19.5 & 29.1 \\

\midrule

BN & N-O & 18.2 & 17.2 & 28.1 & 9.8 & 26.6 & 14.2 & 8.0 & 15.5 & 13.8 & 20.2 & 7.9 & 8.3 & 19.3 & 13.3 & 13.8 & 15.6 \\

Tent & N-O & 16.0 & 14.8 & 24.5 & 9.2 & 23.8 & 13.1 & 7.7 & 14.9 & 13.0 & 16.5 & 8.2 & 8.3 & 17.9 & 10.9 & 13.3 & 14.1 \\

SHOT & N-O & 16.5 & 15.3 & 23.6 & 9.0 & 23.4 & 12.7 & 7.5 & 14.0 & 12.4 & 16.1 & 7.5 & 8.0 & 17.4 & 12.5 & 13.1 & 13.9 \\

\midrule

TTT++ & Y-O & 18.0 & 17.1 & 30.8 & 10.4 & 29.9 & 13.0 & 9.9 & 14.8 & 14.1 & 15.8 & 7.0 & 7.8 & 19.3 & 12.7 & 16.4 & 15.8\\

TTAC & Y-O & 17.9 & 15.8 & 22.5 & \textbf{8.5} & 23.5 & \textbf{11.2} & 7.6 & \textbf{11.9} & 12.9 & \textbf{13.3} & \textbf{6.9} & 7.6 & 17.3 & 12.3 & 12.6 & 13.4 \\

\midrule

TeSLA & N-O & 13.3 & 12.5 & 20.8 & 8.8 & 21.1 & 11.8 & 7.3 & 12.6 & 11.2 & 15.6 & 7.6 & 7.6 & 16.2 & 9.7 & 11.6 & 12.5 \\

TeSLA-s & Y-O & \textbf{13.0} & \textbf{12.2} & \textbf{20.3} & \textbf{8.5} & \textbf{20.8} & \textbf{11.2} & \textbf{7.2} & 12.0 & \textbf{11.0} & 15.5 & 7.3 & \textbf{7.2} & \textbf{15.6} & \textbf{9.1} & \textbf{11.3} & \textbf{12.1} \\

\midrule
\midrule

BN & N-M & 17.3 & 16.2 & 28.0 & 9.8 & 26.1 & 14.0 & 7.9 & 16.1 & 13.7 & 20.4 & 8.3 & 8.3 & 19.6 & 11.8 & 14.0 & 15.4 \\

Tent & N-M & 15.1 & 13.7 & 22.2 & 8.5 & 22.4 & 11.8 & 7.1 & 12.7 & 11.9 & 12.9 & 7.6 & 7.6 & 16.9 & 9.8 & 12.6 & 12.9 \\

SHOT & N-M & 15.8 & 14.8 & 24.9 & 9.2 & 23.6 & 13.2 & 7.5 & 14.5 & 12.8 & 17.5 & 8.1 & 8.2 & 18.1 & 10.8 & 13.4 & 14.2 \\

\midrule

TTT++ & Y-M & 13.2 & 11.8 & \textbf{11.1} & 7.9 & 16.5 & 8.9 & 6.6 & 9.5 & 9.7 & 8.6 & \textbf{5.2} & \textbf{5.6} & 13.1 & 8.8 & 11.1 & 9.8 \\

TTAC & Y-M & 11.6 & 10.3 & 15.8 & \textbf{6.8} & 15.9 & \textbf{7.5} & \textbf{5.8} & \textbf{8.7} & 9.0 & \textbf{8.5} & 5.6 & 5.7 & \textbf{12.7} & 8.0 & 9.7 & \textbf{9.4} \\

\midrule

TeSLA & N-M & 10.7 & \textbf{9.8} & 15.2 & 7.0 & \textbf{15.8} & 9.1 & 6.1 & 10.0 & \textbf{8.9} & 10.9 & 6.0 & 6.2 & 13.0 & \textbf{7.9} & 9.6 & 9.7 \\

TeSLA-s & Y-M & \textbf{10.4} & \textbf{9.8} & 14.9 & 7.3 & 16.1 & 9.0 & 6.2 & 9.5 & 9.1 & 11.5 & 5.9 & 5.8 & 12.9 & \textbf{7.9} & \textbf{9.5} & 9.7 \\

\midrule
\multicolumn{18}{c}{\cellcolor[HTML]{EFEFEF}CIFAR100-C} \\
\midrule


Source & N & 80.8 & 77.8 & 87.8 & 39.6 & 82.3 & 54.2 & 38.4 & 54.6 & 60.2 & 68.1 & 28.9 & 50.9 & 59.5 & 72.3 & 50.0 & 60.4 \\

\midrule

BN & N-O & 48.2 & 46.4 & 61.1 & 33.8 & 58.2 & 41.4 & 31.9 & 46.1 & 42.5 & 54.7 & 31.3 & 33.3 & 48.4 & 39.0 & 39.6 & 43.7\\

Tent & N-O & 43.3 & 41.2 & 52.7 & 31.2 & 50.8 & 36.1 & 29.3 & 41.9 & 38.9 & 43.6 & 30.1 & 31.0 & 43.5 & 34.4 & 36.5 & 39.0 \\

SHOT & N-O & 44.1 & 41.8 & 53.3 & 31.5 & 50.6 & 36.0 & 29.6 & 40.7 & 40.1 & 41.9 & 29.5 & 33.6 & 44.0 & 34.9 & 36.6 & 39.2\\

\midrule

TTT++ & Y-O & 50.2 & 47.7 & 66.1 & 35.8 & 61.0 & 38.7 & 35.0 & 44.6 & 43.8 & 48.6 & 28.8 & 30.8 & 49.9 & 39.2 & 45.5 & 44.4\\

TTAC & Y-O & 47.7 & 45.7 & 58.1 & 32.5 & 55.3 & 36.6 & 31.2 & 40.3 & 40.8 & \textbf{44.7} & 30.0 & 39.9 & 47.1 & 37.8 & 38.3 & 41.7 \\

\midrule

TeSLA & N-O & 40.0 & 38.9 & 51.5 & 32.2 & 49.1 & 36.9 & 29.7 & 40.4 & 37.4 & 46.0 & 29.3 & 30.7 & 42.7 & 32.9 & 34.6 & 38.2 \\

TeSLA-s & Y-O & \textbf{39.1} & \textbf{38.5} & \textbf{50.0} & \textbf{30.6} & \textbf{48.6} & \textbf{35.9} & \textbf{29.1} & \textbf{38.9} & \textbf{36.4} & 46.2 & \textbf{28.3} & \textbf{29.7} & \textbf{41.9} & \textbf{32.1} & \textbf{33.9} & \textbf{37.3} \\

\midrule
\midrule

BN & N-M & 47.4 & 45.5 & 60.0 & 33.9 & 56.9 & 40.8 & 31.8 & 46.4 & 42.6 & 54.2 & 32.3 & 33.1 & 48.5 & 37.2 & 39.4 & 43.3 \\

Tent& N-M & 41.0 & 38.4 & 49.2 & 30.0 & 47.4 & 33.1 & 28.1 & 38.1 & 38.0 & 37.5 & 28.3 & 29.0 & 41.1 & 32.8 & 35.6 & 36.5 \\

SHOT & N-M& 41.6 & 40.6 & 51.7 & 31.4 & 49.5 & 36.2 & 29.3 & 42.4 & 38.4 & 45.4 & 29.9 & 31.3 & 43.1 & 33.5 & 36.0 & 38.7 \\

\midrule

TTT++ & N-M & 38.4 & 37.7 & \textbf{41.3} & 29.1 & 44.1 & 32.9 & 27.8 & 34.3 & 34.4 & \textbf{34.7} & 25.4 & 26.6 & 39.2 &  32.3 & 33.6 & 34.1\\

TTAC & N-M & 37.8 & 36.8 & 45.1 & 28.2 & 45.3 & \textbf{30.7} & 26.6 & 35.3 & 35.7 & 36.7 & 26.8 & 27.4 & 39.6 & 30.6 & 34.2 & 33.6 \\

\midrule

TeSLA & N-M & 34.4 & 33.5 & 42.2 & 28.0 & \textbf{41.9} & 32.1 & \textbf{25.9} & 35.1 & 32.6 & 38.3 & 25.0 & 27.4 & 37.5 & 28.6 & 30.6 & 32.9\\

TeSLA-s & Y-M & \textbf{33.9} & \textbf{33.0} & 42.1 & \textbf{27.5} & 42.0 & 31.6 & 26.1 & \textbf{34.2} & \textbf{32.2} & 39.4 & \textbf{24.8} & \textbf{26.3} & \textbf{36.8} & \textbf{28.1} & \textbf{30.3} & \textbf{32.6} \\

\midrule
\multicolumn{18}{c}{\cellcolor[HTML]{EFEFEF}ImageNet-C} \\
\midrule


Source & N & 97.0 & 96.3 & 97.4 & 82.1 & 90.3 & 85.3 & 77.5 & 83.4 & 76.9 & 76.0 & 40.9 & 94.6 & 83.5 & 79.1 & 67.4 & 81.8 \\

\midrule

BN & N-O & 83.5 & 82.6 & 82.9 & 84.4 & 84.2 & 73.1 & 60.5 & 65.1 & 66.3 & 51.5 & 34.0 & 82.6 & 55.3 & 50.3 & 58.7 & 67.7 \\

Tent & N-O & 70.8 & 68.7 & 69.1 & 72.5 & 73.3 & 59.3 & 50.8 & 53.0 & 59.1 & 42.7 & 32.6 & \textbf{74.5} & 45.5 & 41.6 & 47.8 & 57.4 \\

SHOT & N-O &77.0 & 74.6 & 76.4 & 81.2 & 79.3 & 72.5 & 61.7 & 65.7 & 66.3 & 55.6 & 56.0 & 92.7 & 57.1 & 56.3 & 58.2 & 68.7\\

\midrule

TTAC & Y-O & 71.5& 67.7 &70.3 &81.2 &77.3 &64 &54.4 &51.1 &56.9 &45.4 &32.6 &79.1 &46.0 & 43.7& 48.6& 59.3 \\

TTT++ & Y-O & 69.4 & 66 & 69.7 &84.2 &81.7 &65.2 &53.2 &49.3 &56.2 &44.4 &32.8 &75.7 &43.9 &41.6 &46.9 & 58.7\\

\midrule

TeSLA & N-O & 65.0 & 62.9 & 63.5 & 69.4 & 69.2 & 55.4 & 49.5 & 49.1 & 56.6 & 41.8 & 33.7 & 77.9 & 43.3 & 40.4 & 46.6 & 55.0\\

TeSLA-s & Y-O & \textbf{61.4} & \textbf{58.8} & \textbf{60.3} & \textbf{67.3} & \textbf{66.2} & \textbf{54.0} & \textbf{48.2} & \textbf{46.9} & \textbf{53.1} & \textbf{40.9} & \textbf{32.4} & 81.2 & \textbf{41.1} & \textbf{39.2} & \textbf{44.8} & \textbf{53.1} \\

\midrule
\midrule

BN & N-M & 83.4 & 82.6 & 82.8 & 84.4 & 84.2 & 73.2 & 60.3 & 64.9 & 66.4 & 51.2 & 34.0 & 82.6 & 54.9 & 49.9 & 58.8 & 67.6\\

Tent & N-M & 66.1 & 63.7 & 64.2 & 68.9 & 69.6 & 52.6 & 47.4 & 48.4 & 58.4 & 39.8 & \textbf{31.6} & 77.9 & 41.7 & \textbf{28.7} & 44.5 & 54.2 \\

SHOT & N-M & 75.8 & 73.7 & 73.7 & 78.3 & 77.1 & 71.8 & 60.9 & 64.2 & 66.1 & 55.4 & 59.8 & 95.5 & 56.1 & 57.3 & 58.1 & 68.2\\

\midrule

TeSLA & N-M & \textbf{62.3} & \textbf{60.9} & \textbf{60.6} & \textbf{64.3} & \textbf{65.7} & \textbf{50.4} & \textbf{46.2} & \textbf{46.1} & \textbf{54.7} & \textbf{39.1} & 32.2 & \textbf{68.5} & \textbf{40.9} & 37.5 & \textbf{43.5} & \textbf{51.5}\\

\bottomrule
\end{tabular}}
\label{tab:cifar_c}
\end{table*}

\begin{table*}[htb]
\centering
\caption{\textbf{Comparison of state-of-the-art TTA methods under different protocols} on the VisDA-C dataset. We report the class top-1 accuracies (\%) for each of the 12 classes. We also report the overall accuracy (Acc.) and the per-class average accuracy (Avg.). Each result is averaged over three seeds.}
\resizebox{\linewidth}{!}{
\begin{tabular}{@{}c|c|cccccccccccc|cc}
\toprule
     Method & \rotatebox[origin=c]{90}{Protocol} & plane & bicycle & bus & car & horse & knife & mcycl & person & plant & sktbrd & train & truck & Acc. & Avg. \\
     
\midrule

Source & N & 3.8\std{4.5} & 23.3\std{0.7} & 56.0\std{3.9} & 82.5\std{0.9} & 70.8\std{3.1} & 1.6\std{0.3} & 84.4\std{1.3} & 9.1\std{2.2} & 78.0\std{5.4} & 22.1\std{3.7} & 79.3\std{2.4} & 1.6\std{0.7} & 55.6\std{0.7} & 48.5\std{1.0} \\

\midrule

BN & N-O & 86.9\std{2.2} & 57.8\std{2.3} & 75.4\std{1.2} & 52.9\std{1.3} & 86.7\std{0.6} & 54.2\std{4.0} & 85.5\std{0.9} & 55.4\std{2.0} & 64.9\std{2.7} & 41.6\std{2.3} & 85.7\std{1.2} & 28.8\std{2.5} & 64.5\std{0.3} & 64.6\std{0.5} \\

Tent & N-O  & 86.9\std{2.2} & 57.7\std{3.0} & 77.4\std{1.4} & 56.8\std{1.5} & 87.3\std{0.8} & 62.4\std{3.8} & 86.6\std{0.8} & 62.9\std{2.9} & 71.2\std{1.7} & 39.9\std{2.8} & 84.8\std{1.2} & 24.7\std{3.4} & 66.3\std{0.3} & 66.5\std{0.6} 
\\

SHOT & N-O  & 90.5\std{1.0} & 77.0\std{0.9} & 76.2\std{0.7} & 47.5\std{0.5} & 87.9\std{0.2} & 62.1\std{4.0} & 75.9\std{0.2} & 74.4\std{1.1} & 83.3\std{0.3} & 47.0\std{6.6} & 84.2\std{0.9} & 41.6\std{0.4} & 68.6\std{0.6} & 70.6\std{1.0} 
 \\
 
AdaContrast & N-O  & 95.2\std{0.3} & 78.2\std{0.3} & 81.8\std{0.1} & 67.9\std{1.2} & 94.9\std{0.5} & 87.4\std{3.3} & \textbf{87.9\std{0.6}} & 82.0\std{1.5} & 90.7\std{0.7} & 36.8\std{16.1} & \textbf{88.6\std{0.1}} & 31.5\std{3.6} & 76.2\std{0.7} & 76.9\std{1.4} \\

\midrule

TTT++ & Y-O & 86.4\std{1.5} & 60.5\std{2.6} & 75.7\std{2.2} & 51.7\std{3.6} & 86.5\std{0.9} & 55.3\std{2.1} & 85.2\std{2.7} & 55.8\std{1.1} & 64.5\std{2.7} & 41.3\std{2.1} & 86.4\std{1.9} & 28.4\std{2.6} & 64.4\std{0.8} & 64.8\std{0.7} \\

TTAC & Y-O & 90.0\std{1.2} & 64.7\std{12.5} & 69.7\std{0.9} & 48.5\std{1.7} & 84.3\std{1.8} & 82.8\std{3.6} & 84.7\std{4.1} & 64.7\std{7.2} & 72.1\std{1.3} & 40.2\std{6.3} & 86.5\std{1.2} & 25.5\std{5.6} & 65.5\std{1.6} & 67.8\std{2.1} \\

\midrule

TeSLA & N-O  & \textbf{95.4\std{0.2}} & \textbf{87.4\std{0.2}} & \textbf{83.8\std{0.6}} & \textbf{70.1\std{0.8}} & \textbf{95.1\std{0.1}} & 90.0\std{1.0} & 84.8\std{3.1} & \textbf{83.2\std{1.3}} & \textbf{93.6\std{0.1}} & \textbf{67.9\std{19.9}} & 85.4\std{0.8} & \textbf{49.3\std{1.2}} & \textbf{80.3\std{1.3}} & \textbf{82.2\std{1.9}} \\

TeSLA-s & Y-O & 92.0\std{0.2} & 81.2\std{2.0} & 77.1\std{1.9} & 56.5\std{0.9} & 90.2\std{0.4} & \textbf{91.0\std{0.9}} & 82.9\std{1.8} & 79.8\std{0.8} & 91.3\std{0.1} & 48.9\std{3.5} & 81.2\std{1.5} & 40.1\std{2.4} & 73.5\std{0.3} & 76.0\std{0.3} \\

\midrule
\midrule

BN & N-M  & 87.2\std{1.4} & 58.0\std{1.1} & 76.4\std{1.4} & 53.7\std{1.9} & 87.2\std{1.3} & 54.2\std{3.6} & 86.2\std{0.3} & 55.5\std{1.6} & 64.9\std{2.3} & 42.1\std{2.7} & 85.6\std{1.3} & 29.3\std{2.2} & 64.9\std{0.1} & 65.0\std{0.4}\\

Tent & N-M   & 89.1\std{2.0} & 56.4\std{5.9} & 82.4\std{1.0} & 69.2\std{0.5} & 89.3\std{1.3} & 95.2\std{0.5} & \textbf{91.4\std{0.5}} & 79.5\std{1.0} & 86.1\std{0.3} & 16.3\std{1.9} & 84.7\std{0.4} & 8.4\std{3.5} & 70.9\std{0.4} & 70.7\std{0.6}  \\

SHOT & N-M   & 93.9\std{0.5} & 82.6\std{0.7} & 76.6\std{0.8} & 49.7\std{1.8} & 92.0\std{0.2} & 79.0\std{21.6} & 75.3\std{2.0} & 80.9\std{2.4} & 89.5\std{0.6} & 50.5\std{19.0} & 83.8\std{0.9} & 52.2\std{1.1} & 72.7\std{1.8} & 75.5\std{3.4} \\

AdaContrast & N-M   & 95.6\std{0.6} & 82.8\std{1.0} & 76.5\std{2.4} & \textbf{72.4\std{5.3}} & \textbf{96.7\std{0.3}} & 91.3\std{2.2} & 88.6\std{1.2} & \textbf{85.4\std{0.8}} & 95.3\std{0.5} & 30.1\std{51.3} & \textbf{93.6\std{0.7}} & 48.9\std{2.1} & 79.7\std{1.3} & 79.8\std{3.9} \\

\midrule

TTT++ & Y-M & 87.2\std{2.0} & 61.8\std{2.0} & 74.7\std{1.3} & 52.7\std{3.6} & 86.1\std{1.7} & 65.0\std{7.0} & 84.9\std{2.3} & 62.1\std{6.0} & 67.2\std{1.6} & 36.6\std{1.3} & 86.2\std{0.1} & 27.1\std{3.6} & 65.3\std{0.3} & 65.9\std{1.0}\\

TTAC & Y-M &  86.8\std{4.2} & 73.5\std{1.3} & 69.3\std{2.1} & 44.2\std{2.5} & 78.8\std{5.1} & 73.1\std{6.7} & 84.7\std{1.6} & 67.3\std{8.6} & 78.6\std{5.6} & 52.9\std{4.1} & 84.7\std{2.6} & 33.2\std{3.9} & 66.0\std{2.0} & 68.9\std{2.4}\\

\midrule

TeSLA & N-M   & \textbf{96.6\std{0.2}} & \textbf{91.3\std{0.1}} & \textbf{85.1\std{1.0}} & 69.3\std{0.0} & \textbf{96.7\std{0.3}} & \textbf{97.1\std{0.8}} & 88.0\std{0.9} & 85.2\std{0.4} & \textbf{96.3\std{0.2}} & \textbf{87.7\std{9.3}} & 87.4\std{0.2} & \textbf{57.3\std{0.8}} & \textbf{83.4\std{0.6}} & \textbf{86.5\std{0.9}} \\

TeSLA-s & Y-M & 96.1\std{0.4} & 89.4\std{0.4} & 83.0\std{0.4} & 62.4\std{0.5} & 94.4\std{0.1} & 94.5\std{1.1} & 87.3\std{0.3} & 83.3\std{0.5} & 95.5\std{0.2} & 63.9\std{14.9} & 85.7\std{0.7} & 49.4\std{3.3} & 79.3\std{1.0} & 82.1\std{1.5} \\

\bottomrule
\end{tabular}}
\label{tab:visda}
\end{table*}


\begin{table*}[htb]
\caption{\textbf{Segmentation results} for test-time adaptation methods (class Avg. volume-wise mean Dice score in \%) on the spinal cord dataset (site $\left \{\texttt{1}  \right \}\rightarrow \texttt{2,3,4}$) and prostate dataset (sites $\left \{ \texttt{A,B} \right \}\rightarrow \texttt{D,E,F}$), respectively.}
\resizebox{\linewidth}{!}{%
\begin{tabular}{@{}c|c|ccc|c||c|c|c|c@{}}
\toprule
Method              & \rotatebox[origin=c]{0}{Protocol}             & \multicolumn{4}{c||}{Spinal Cord}                                                                                    & \multicolumn{4}{c}{Prostate}                                  \\
\midrule
\multicolumn{2}{c|}{\multirow{2}{*}{Sites}} & $\left \{ \texttt{1} \right \}\rightarrow\left \{\texttt{2}  \right \}$ & $\left \{ \texttt{1} \right \}\rightarrow\left \{\texttt{3}  \right \}$ & $\left \{ \texttt{1} \right \}\rightarrow\left \{\texttt{4}  \right \}$ & $\left \{ \texttt{1} \right \}\rightarrow\left \{\texttt{2,3,4}  \right \}$ & $\left \{ \texttt{A,B} \right \}\rightarrow\left \{\texttt{D}  \right \}$ & $\left \{ \texttt{A,B} \right \}\rightarrow\left \{\texttt{E}  \right \}$ & $\left \{ \texttt{A,B} \right \}\rightarrow\left \{\texttt{F}  \right \}$ & $\left \{ \texttt{A,B} \right \}\rightarrow\left \{\texttt{D,E,F}  \right \}$ \\
\multicolumn{2}{c|}{}  & Class Avg.      & Class Avg.  & Class Avg.     &   Avg.   & Class Avg.         & Class Avg.         & Class Avg.         &   Avg.   \\

\midrule
Source & N  & 77.4\std{6.6} & 64.8\std{11.7} & 85.9\std{3.8} & 76.0 \std{11.8} & 75.8\std{8.9} & 65.9\std{18.5} & 38.4\std{32.3} & 60.5\std{27.0} \\

\midrule

BN     & N-O  & 85.2\std{2.1}  & 70.6\std{3.6}  & 88.9\std{1.7} & 81.6\std{8.3} & 75.9\std{9.4} & 74.4\std{7.4} & 65.7\std{22.4} & 72.1\std{15.2} \\
TENT   & N-O  & 85.7\std{1.8}  & 68.7\std{2.8}  & 88.9\std{1.7} & 81.1\std{9.1} & 78.8\std{6.2} & 77.9\std{6.9} & 67.0\std{28.4} & 74.7\std{17.9}\\
PL     & N-O  & 85.3\std{2.1}  & 71.0\std{3.6}  & 88.9\std{1.7} & 81.7\std{8.6} & 76.1\std{9.4} & 74.8\std{7.5} & 66.2\std{22.4} & 72.4\std{15.2}  \\
OptTTA & N-O  & 84.4\std{2.3}  &  80.2\std{5.1}  & 87.5\std{2.0}  & 84.1\std{4.8}     & 84.9\std{6.9} & \textbf{80.3\std{8.4}} & 84.0\std{6.6} & 83.1\std{7.7}\\
TeSLA  & N-O  & \textbf{86.3\std{1.5}}  & \textbf{80.3\std{7.3}} & \textbf{89.3\std{1.4}} & \textbf{85.3\std{5.8}} & \textbf{86.1\std{3.3}} &  79.8\std{7.5} & \textbf{84.3\std{6.3}} & \textbf{83.5\std{6.5}} \\
\midrule \midrule
BN  & N-M & 85.5\std{1.6} & 78.5\std{3.2} & 88.8\std{1.5} & 84.3\std{4.8} & 77.8\std{9.6} & 77.3\std{7.2} & 63.8\std{26.7} & 73.1\std{18.0}  \\
TENT& N-M  & 85.5\std{1.6} & 79.0\std{3.3} & 88.8\std{1.5} & 84.4\std{4.7} & 81.6\std{7.7} & 79.0\std{10.4} & 82.8\std{9.2} & 81.2\std{9.3} \\
PL  & N-M  & 85.5\std{1.7} & 78.8\std{3.3} & 88.8\std{1.5} & 84.3\std{4.7} & 81.2\std{7.9} & 79.1\std{10.1} & 82.8\std{9.1} & 81.1\std{9.2}\\
OptTTA & N-M  & 84.3\std{2.5} & \textbf{80.7\std{4.9}} & 87.7\std{2.0} & 84.3\std{4.4}  & \textbf{86.2\std{5.2}} & 78.6\std{8.6} & 85.0\std{6.7} & 83.4\std{7.7}\\
TeSLA  & N-M  & \textbf{86.4\std{1.7}}  & 80.4\std{3.2} & \textbf{89.3\std{1.7}} & \textbf{85.4\std{4.4}} & 85.9\std{4.0} & \textbf{81.2\std{6.7}} & \textbf{85.6\std{5.4}} & \textbf{84.3\std{5.8}} \\
\bottomrule
\end{tabular}%
}
\label{tab:mri}
\end{table*}
\begin{table*}[t!]
\centering
\caption{\textbf{Segmentation results} for test-time adaptation methods (mIoU\%)  for adaptation from synthetic GTA5 dataset to Cityscapes dataset (O and M protocols). }
\resizebox{\linewidth}{!}{
\begin{tabular}{@{}c|c|ccccccccccccccccccc|c@{}}
\toprule
Method & \begin{sideways}Protocol\end{sideways} & \begin{sideways}road\end{sideways} & \begin{sideways}sidewalk\end{sideways} & \begin{sideways}building\end{sideways} & \begin{sideways}wall\end{sideways} & \begin{sideways}fence\end{sideways} & \begin{sideways}pole\end{sideways} & \begin{sideways}light\end{sideways} & \begin{sideways}sign\end{sideways} & \begin{sideways}vegetation\end{sideways} & \begin{sideways}terrain\end{sideways} & \begin{sideways}sky\end{sideways} & \begin{sideways}person\end{sideways} & \begin{sideways}rider\end{sideways} & \begin{sideways}car\end{sideways} & \begin{sideways}truck\end{sideways} & \begin{sideways}bus\end{sideways} & \begin{sideways}train\end{sideways} & \begin{sideways}motocycle\end{sideways} & \begin{sideways}bicycle\end{sideways} & mIoU \\
\midrule
Source & N & 72.9 & 20.0 & 81.4 & 21.7 & 22.9 & 19.2 & 25.3 & 10.6 & 78.9 & 26.4 & 85.9 & 54.9 & 20.7 & 53.0 & 30.6 & 16.2 &  1.9 & 20.0 &  7.5 & 35.3 \\
\midrule
BN & N-O & 84.3 & 31.8 & 79.2 & 24.1 & 20.5 & 21.5 & 23.5 & 10.5 & 74.6 & 32.0 & 75.3 & 52.1 & 14.4 & 77.6 & 28.1 & 20.6 &  6.0 & 14.9 &  6.2 & 36.7\\
PL & N-O & 84.0 & 31.1 & 80.6 & 25.5 & 20.7 & 21.5 & 24.7 & 11.4 & 77.3 & 34.0 & \textbf{79.4} & 54.2 & 17.1 & 78.3 & 30.1 & 21.7 &  9.7 & 18.5 &  8.1 & 38.3\\
Tent & N-O & 88.0 & 34.3 & 80.7 & 27.7 & 17.8 & 19.3 & 22.1 & 10.0 & \textbf{80.1} & \textbf{40.5} & 77.6 & 51.8 & 15.7 & 81.8 & \textbf{32.6} & 24.0 &  8.9 & 18.8 &  5.8 & 38.8\\
CoTTA  & N-O &  85.8 & 35.3 & 79.1 & 26.5 & 20.3 & 19.8 & 21.7 &  9.9 & 76.7 & 36.2 & 74.6 & 53.2 & 14.4 & 77.8 & 29.0 & 19.3 &  3.6 & 13.2 &  5.8 & 37.0 \\
TeSLA & N-O &  \textbf{90.4} & \textbf{52.2} & \textbf{82.5} & \textbf{29.6} & \textbf{25.5} & \textbf{28.1} & \textbf{32.5} & \textbf{29.7} & 79.7 & 39.0 & 75.2 & \textbf{59.0} & \textbf{21.3} & \textbf{84.0} & 29.3 & \textbf{24.6} & \textbf{14.5} & \textbf{23.0} & \textbf{26.1} & \textbf{44.5}\\
\midrule
\midrule
BN & N-M & 84.3 & 31.1 & 80.7 & 25.4 & 21.0 & 22.6 & 25.6 & 11.8 & 76.7 & 32.7 & 77.6 & 54.8 & 17.2 & 79.7 & 29.7 & 21.7 &  9.3 & 18.7 &  8.5 & 38.4\\
PL & N-M & 85.1 & 30.6 & 80.9 & 25.7 & 20.6 & 21.5 & 24.7 & 11.3 & 77.9 & 33.9 & \textbf{80.2} & 54.4 & 17.4 & 80.0 & 30.0 & 21.9 &  9.2 & 19.0 &  8.3 & 38.6\\
Tent & N-M & 89.0 & 35.1 & 81.0 & 28.4 & 17.0 & 19.5 & 22.9 &  9.8 & \textbf{80.8} & \textbf{41.8} & 76.7 & 52.4 & 16.3 & 83.6 & \textbf{33.0} & 24.8 &  7.1 & 20.6 &  5.7 & 39.2 \\
CoTTA & N-M & 88.6 & 40.2 & 80.6 & \textbf{30.0} & 20.4 & 19.2 & 25.9 & 16.0 & 77.1 & 32.7 & 75.3 & 55.7 & 23.1 & 82.8 & 30.1 & 19.4 &  9.8 & 19.9 & 11.3 & 39.9 \\
TeSLA & N-M & \textbf{90.1} & \textbf{51.4} & \textbf{83.1} & 29.0 & \textbf{27.7} & \textbf{28.7} & \textbf{34.8} & \textbf{34.0} & 78.7 & 35.7 & 73.0 & \textbf{62.0} & \textbf{26.5} & \textbf{83.9} & 28.5 & \textbf{25.0} & \textbf{25.7} & \textbf{27.3} & \textbf{29.4} & \textbf{46.0} \\

\bottomrule
\end{tabular}}
\label{tab:visdas}
\end{table*}

    
    
    
    
    
    
    
    

    
    
    
    
    
    
     
    

\section{Runtime Analysis}\label{appendix:runtime}
\renewcommand\thefigure{\thesection.\arabic{figure}}
\renewcommand\thetable{\thesection.\arabic{table}}
\setcounter{figure}{0}
\setcounter{table}{0}
We compare the computational runtime cost of several test-time adaptation methods, including BN \cite{ioffe2015batch, nado2020evaluating}, TTAC \cite{su2022revisiting}, SHOT-IM \cite{liang2020we}, TENT \cite{wang2021tent}, AdaContrast \cite{chen2022contrastive} and our proposed method TeSLA in Table \ref{tab:computation_efficiency}. We also include overall TesLA runtimes using static, pre-optimized RandAugment (RA) / AutoAugment (AA) augmentation policies instead of the proposed Adversarial Augmentations.

\begin{table}[H]
    \centering
    \caption{Runtime (GPU hours) per epoch on GeForce RTX-3090 for ResNet-101 with batch size of 128 on the VisDA-C. RA implies RandAugment [10], AA implies AutoAugment [9].}
    \resizebox{\linewidth}{!}{
    \begin{tabular}{@{}ccccc|ccc@{}}
    \midrule
         BN & TTAC & SHOT-IM & TENT & AdaContrast & \textbf{TeSLA} & \textbf{TeSLA (RA)} & \textbf{TeSLA (AA)} \\
    \midrule
        
        0.04 & 0.14 & 0.16 & 0.05 & 0.22 & \cellcolor{cyan!20}\textbf{0.38} & \cellcolor{cyan!20}\textbf{0.27} & \cellcolor{cyan!20}\textbf{0.28} \\
    \midrule
    \end{tabular}}
    \label{tab:computation_efficiency}
\end{table}

\section{Equivalence to other Test-Time Objectives}\label{appendix:other_tt_obj}
\renewcommand\thefigure{\thesection.\arabic{figure}}
\renewcommand\thetable{\thesection.\arabic{table}}
\setcounter{figure}{0}
\setcounter{table}{0}
Our proposed flipped cross-entropy loss \textit{f-}$\mathbb{CE}$ of Eq. \ref{eq1} without soft-pseudo labels from the teacher is equivalent to entropy minimization of TENT \cite{wang2021tent}, while our final objective $\mathcal{L}_\text{TeSLA}$ of Eq. \ref{eq:loss_TeSLA} without the knowledge distillation from adversarial augmentation is equivalent to SHOT-IM \cite{liang2020we} as
$\mathcal{D}_{\text{KL}}\left (\mathbf{Y}\parallel\hat{\mathbf{Y}}\mid \mathbf{X}\right ) = 0$ when the teacher network is an instant update of the student (momentum $\alpha$ is 0). Thus, without the mean-teacher and adversarial augmentation, our method would have similar shortcomings as that of TENT and SHOT. Incorporating the soft-pseudo labels from the mean teacher alone improves TeSLA's accuracy on VisDA-C from 82.0\% to 86.5\%.

\section{Sensitivity Tests and Aditional Ablations}\label{appendix:sensitivity}
\renewcommand\thefigure{\thesection.\arabic{figure}}
\renewcommand\thetable{\thesection.\arabic{table}}
\setcounter{figure}{0}
\setcounter{table}{0}
\subsection{Sensitivity Tests}
\begin{figure}[t]
    \centering
    \begin{subfigure}{0.48\linewidth}
        \includegraphics[width=\linewidth]{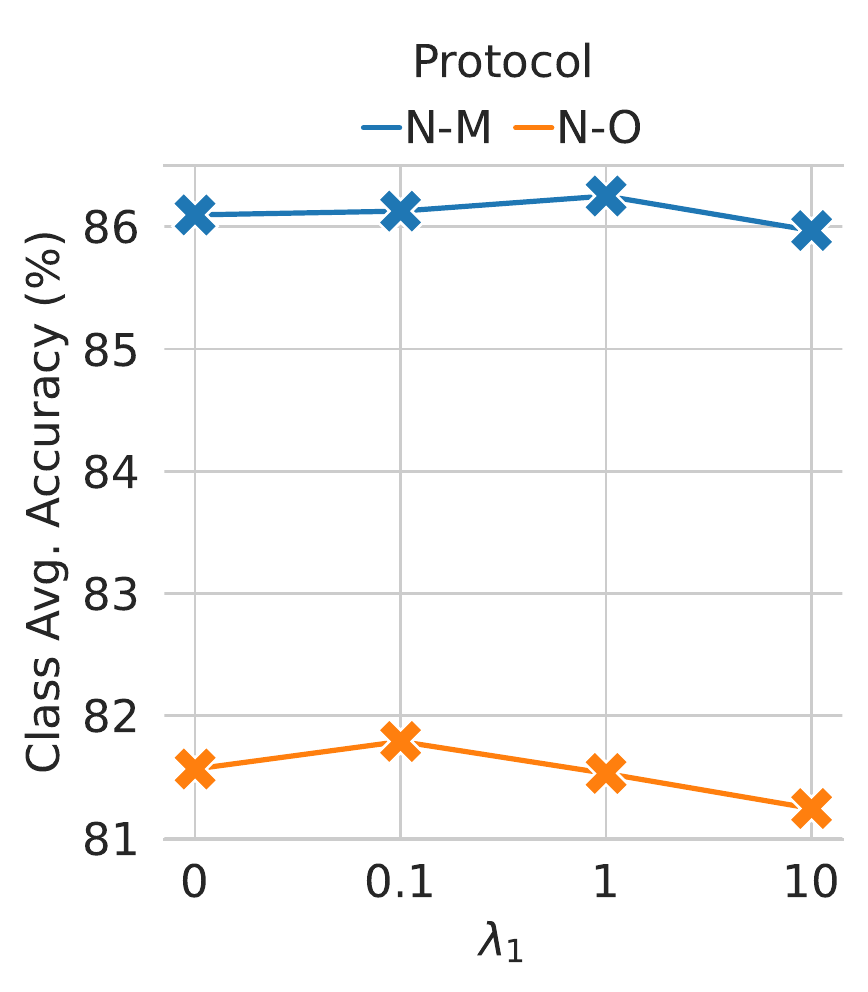}
        \caption{}
    \end{subfigure}
    \begin{subfigure}{0.48\linewidth}
        \includegraphics[width=\linewidth]{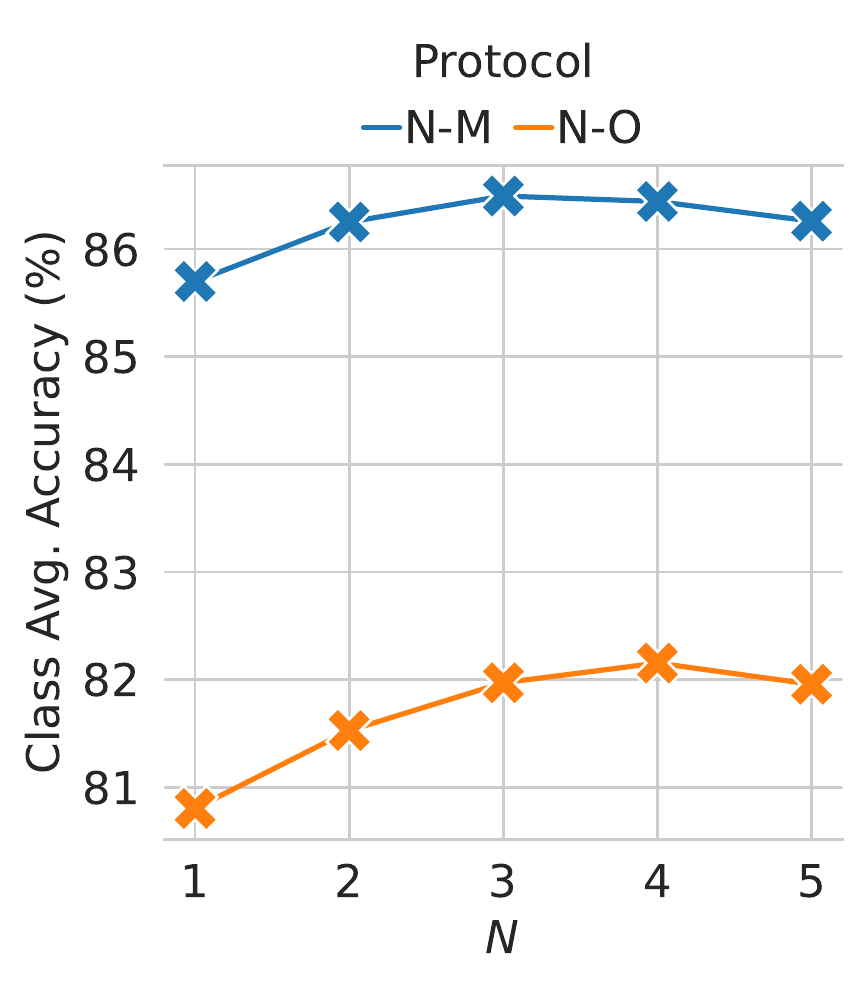}
        \caption{}
    \end{subfigure}
    \caption{\textbf{Sensitivity test for adversarial augmentation hyperparameters} of TeSLA on the VisDA-C dataset for classification task on various TTA protocols. We plot the class Avg. accuracy (\%) on the VisDA-C dataset for (a) augmentation severity controller $\lambda_1 \in \{0, 0.1, 1, 10\}$ and (b) sub-policy dimension $N \in [1,5]$.}
    \label{fig:ablation_augmentation}
\end{figure}

\begin{figure}[h!]
    \centering
    \begin{subfigure}{0.48\linewidth}
        \includegraphics[width=\linewidth]{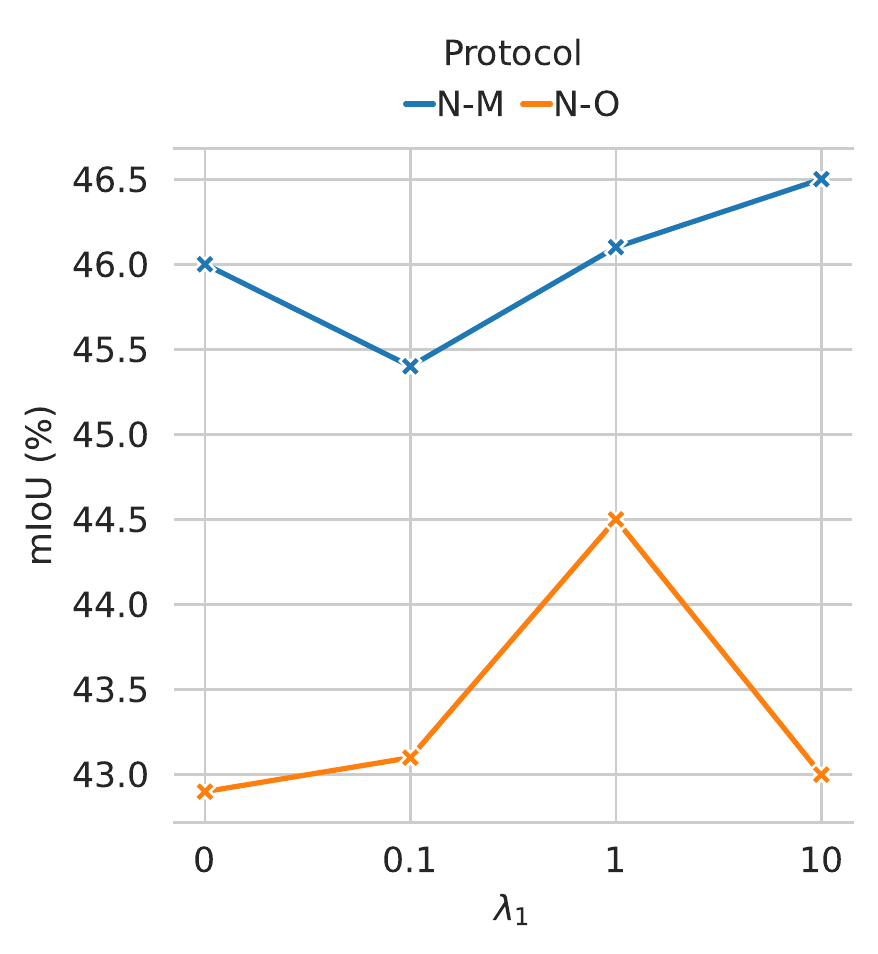}
        \caption{}
    \end{subfigure}
    \begin{subfigure}{0.48\linewidth}
        \includegraphics[width=\linewidth]{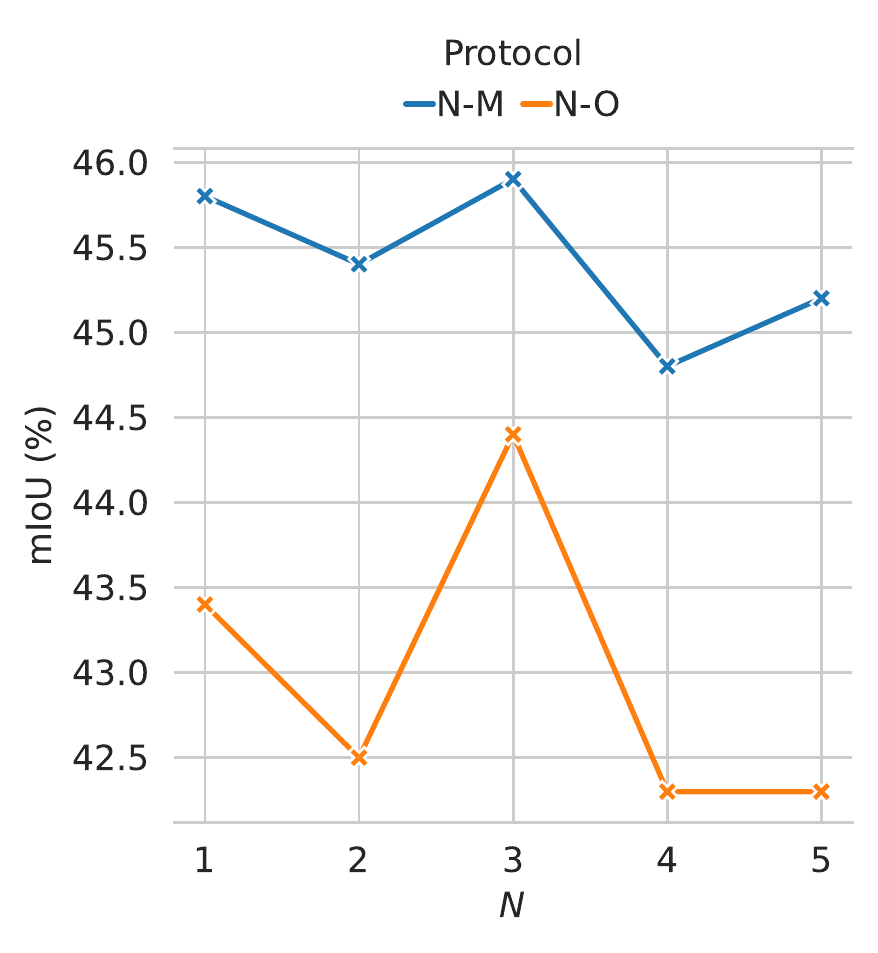}
        \caption{}
    \end{subfigure}
    \caption{\textbf{Sensitivity test for adversarial augmentation hyperparameters} of TeSLA on the VisDA-S dataset for segmentation task on various TTA protocols. We plot the mIoU (\%) on the VisDA-S dataset for (a) augmentation severity controller $\lambda_1 \in \{0, 0.1, 1, 10\}$ and (b) sub-policy dimension $N \in [1,5]$, respectively.}
    \label{fig:ablation_augmentation_seg}
\end{figure}

\paragraph{Automatic Adversarial Augmentation.}
We additionally provide the sensitivity test results for the hyperparameters of the automatic adversarial augmentation module ($\lambda_1$ and sub-policy dimension $N$) on the VisDA-C and VisDA-S datasets. In Fig. \ref{fig:ablation_augmentation}, we show how the class average (Avg.) accuracy (\%) varies with the hyperparameter $\lambda_1$ controlling the severity of augmentations and the sub-policy dimension $N$ on the VisDA-C dataset. Similarly, Fig. \ref{fig:ablation_augmentation_seg} shows the effect of changing $\lambda_1$ and $N$ on the segmentation scores measured by mIoU (\%) on the VisDA-S dataset. We observe that the performance of our method TeSLA is stable over a wide range of $\lambda_1$ and $N$ for both classification and segmentation tasks.

\begin{figure}
    \centering
    \begin{subfigure}{0.48\linewidth}
        \includegraphics[width=\linewidth]{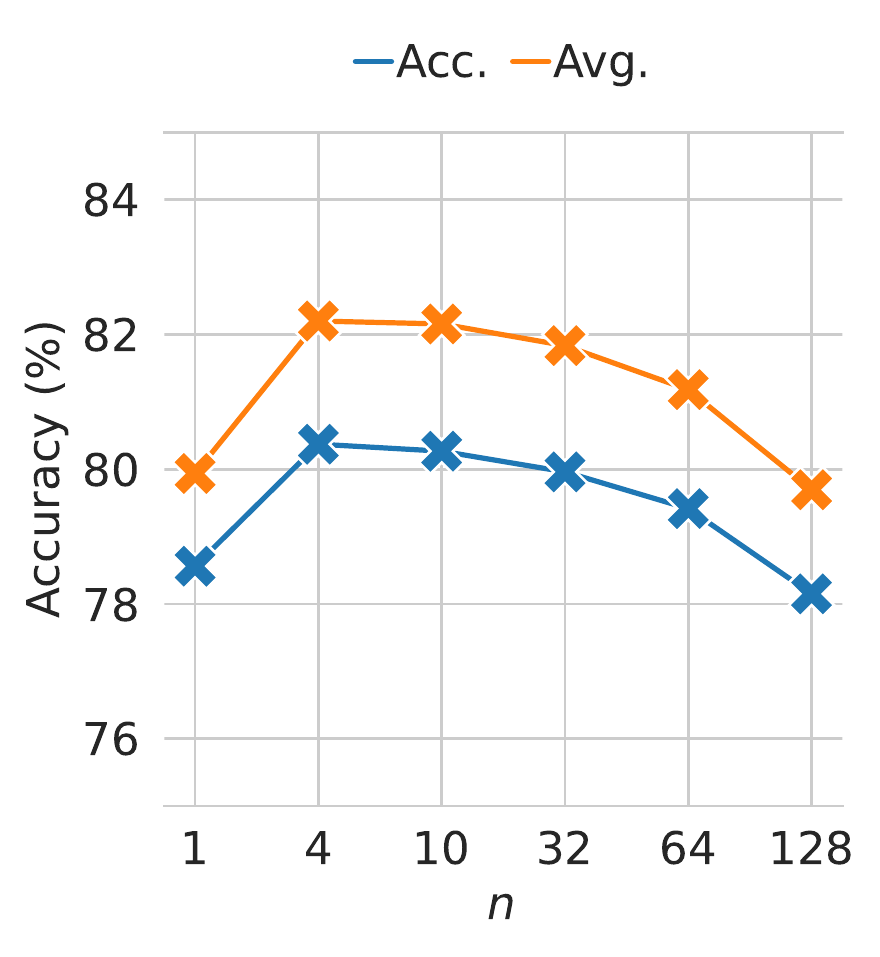}
        \caption{}
    \end{subfigure}
    \begin{subfigure}{0.48\linewidth}
        \includegraphics[width=\linewidth]{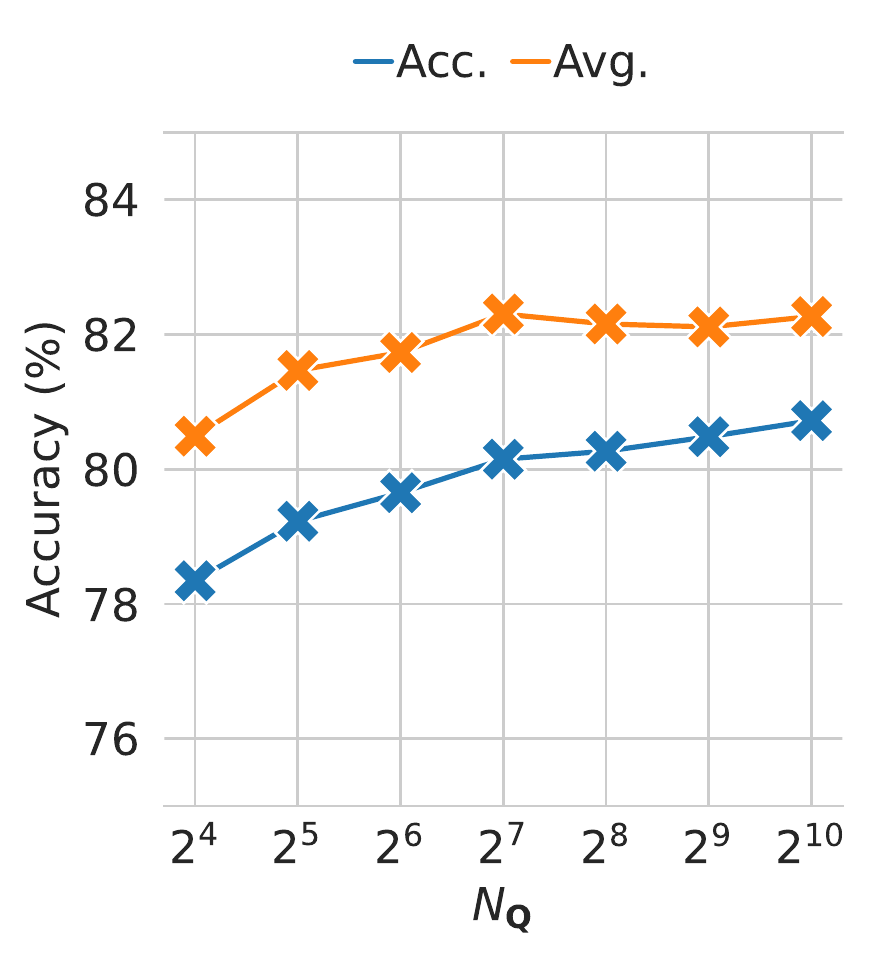}
        \caption{}
    \end{subfigure}
    \caption{\textbf{Sensitivity test for PLR hyperparameters}: (a) the number of nearest neighbors $n$ and (b) the class memory queue size $N_\mathbf{Q}$ on the VisDA-C dataset. We report the overall accuracy (Acc.) and the class average accuracy (Avg.) in \% on the ViSDA-C under the N-O protocol.}
    \label{fig:ablation_nn}
\end{figure}

\paragraph{PLR hyperparameters.}
We present sensitivity tests for the hyperparameters of the soft pseudo-label refinement (PLR) module. In Fig. \ref{fig:ablation_nn},  we show the test-time adaptation classification performance of TeSLA on the VisDA-C (N-O) for varying numbers of nearest neighbors $n\in\{1, 4, 10, 32, 64, 128\}$, and class memory queue size $N_\mathbf{Q}\in\{16, 32, 64, 128, 256, 512, 1024\}$. TeSLA outperforms competing baselines under a wide range of choices. Moreover, the number of examples in the queue can be as small as less than 0.5\% of the dataset size and still maintains on-par performance.

\begin{figure}
\centering
    \begin{subfigure}{0.49\linewidth}
        \includegraphics[width=0.905\linewidth]{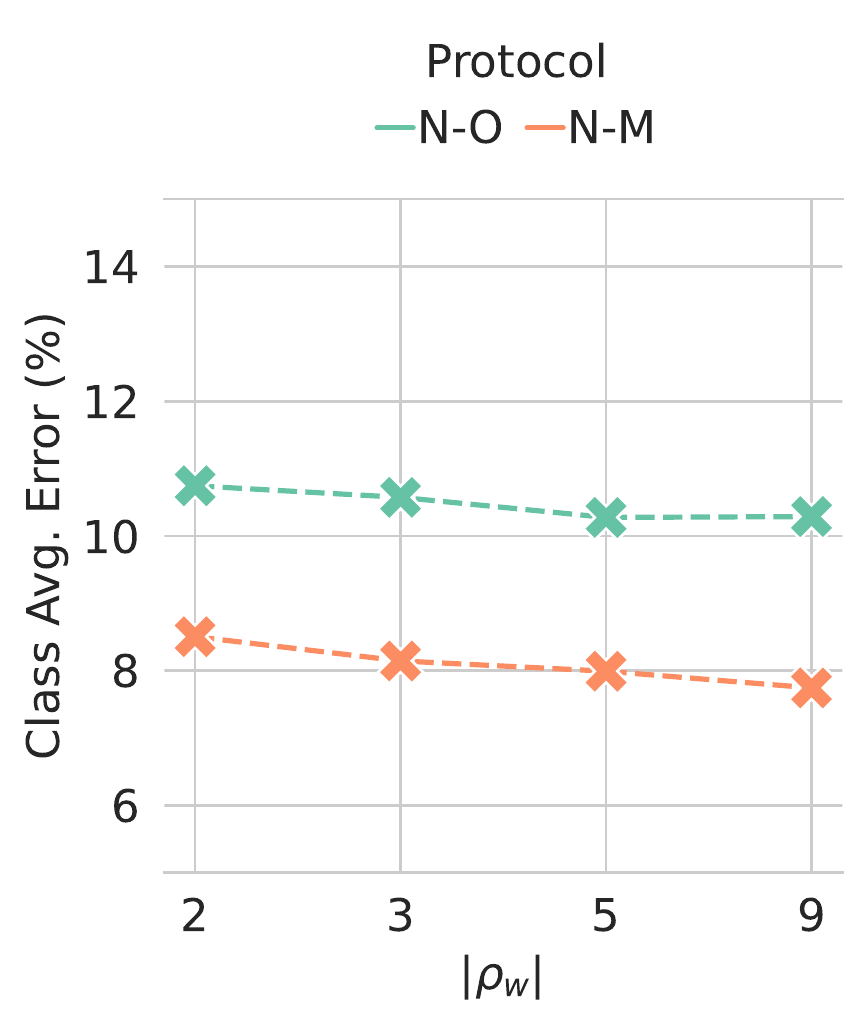}
        \caption{CIFAR10-C}
    \end{subfigure}
    \begin{subfigure}{0.49\linewidth}
        \includegraphics[width=0.905\linewidth]{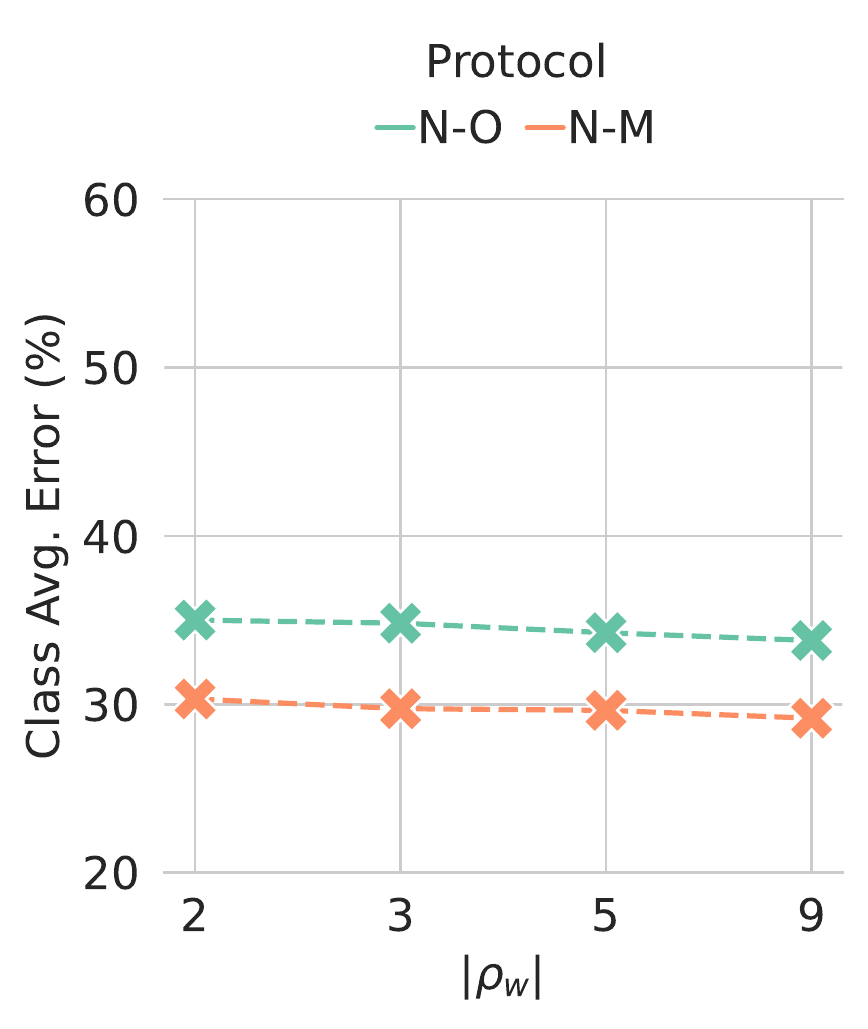}
        \caption{CIFAR100-C}
    \end{subfigure}
    \caption{\textbf{Sensitivity test on the number of weak augmentation} ($|\rho_w|$) for ensembling soft-pseudo labels for Soft-Pseudo Label Refinement (PLR) on (a) CIFAR10-C and (b) CIFAR100-C datasets. We report, for each case, the average error rate over 4 validation corruptions.}
    \label{fig:views_ablation_cifar100}
\end{figure}

Fig. \ref{fig:views_ablation_cifar100} shows the classification performance of TeSLA on the CIFAR10-C and CIFAR100-C and various corruptions [{\footnotesize\textbf{\textsc{Gaussian Blur, Spatter, Speckle Noise, Saturate}}}] under multiple protocols and the number of weak augmentation for ensembling $|\rho_w|\in\{2,3,5,9\}$. These plots show that increasing the number of views decreases the average error rate in both protocols. While we report the results with $|\rho_w|=5$ in the main, we observe we could further decrease the error rate with $|\rho_w|=9$. However, this choice would multiply the computational cost by two as the number of forward passes is linearly proportional to this hyperparameter. For this reason, we opt $|\rho_w|=5$, which gathers the benefit of ensembling and reasonable computational cost.

\begin{figure}
\centering
\begin{subfigure}{0.48\linewidth}
    \includegraphics[width=\linewidth]{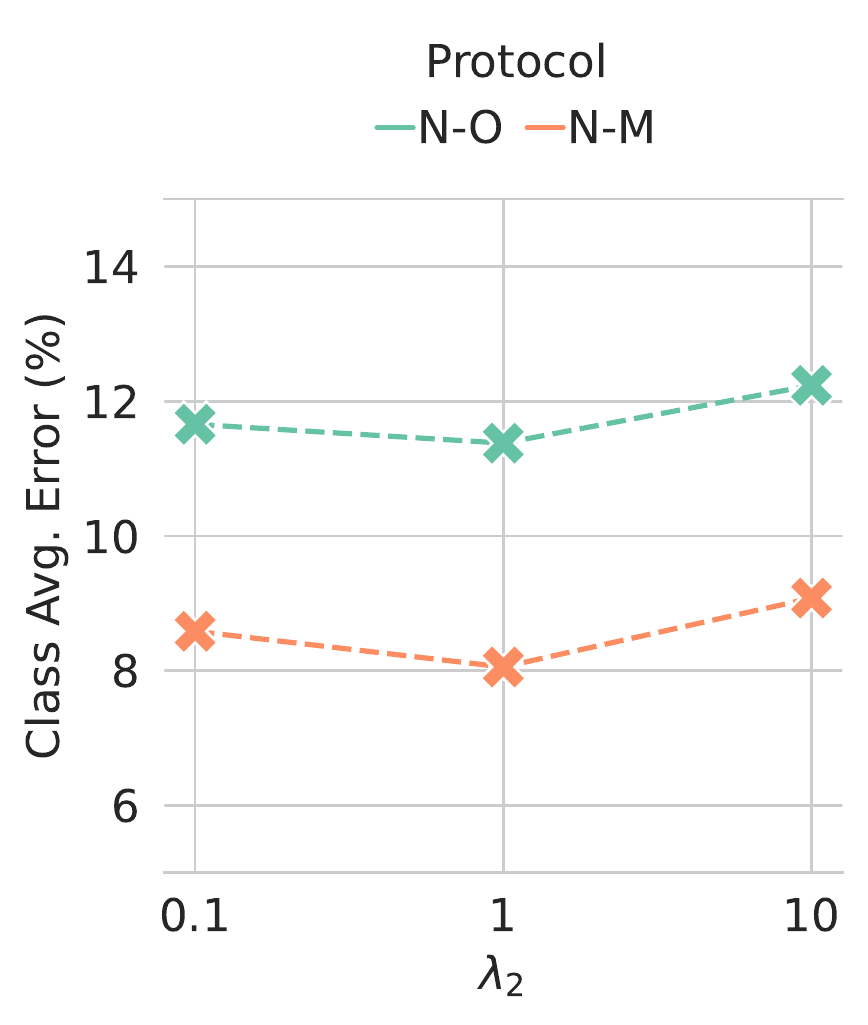}
    \caption{CIFAR10-C}
\end{subfigure}
\begin{subfigure}{0.48\linewidth}
    \includegraphics[width=\linewidth]{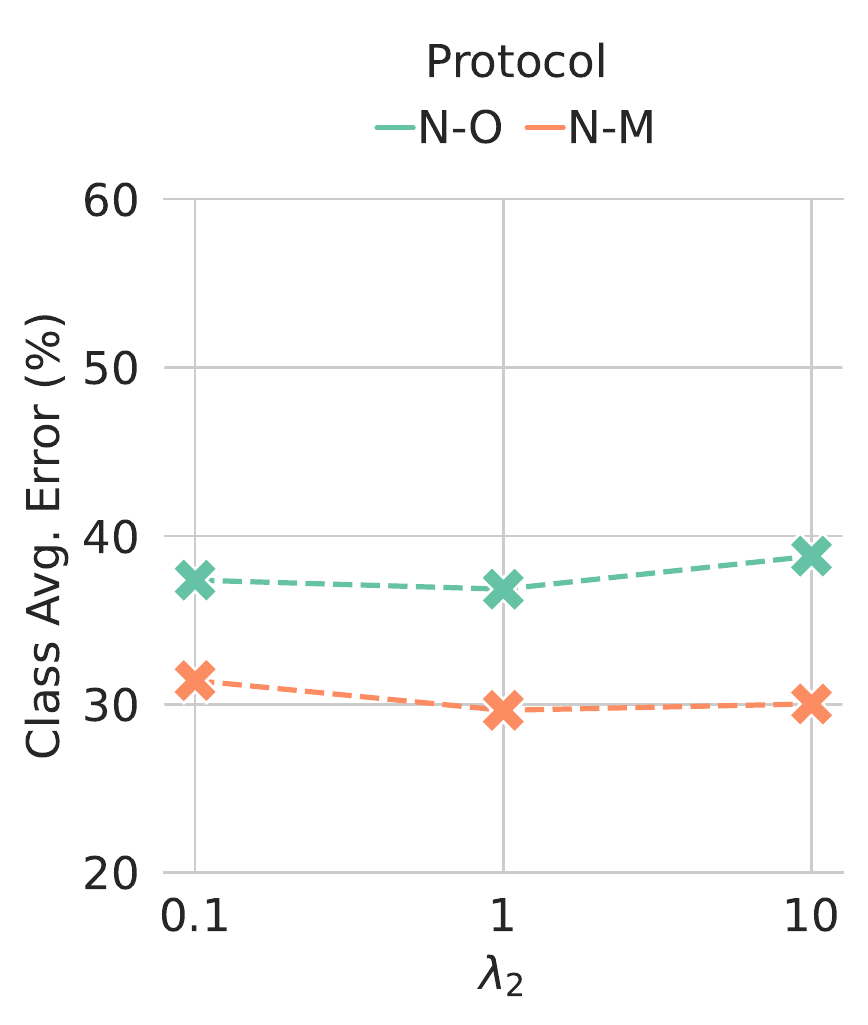}
    \caption{CIFAR100-C}
\end{subfigure}
    \caption{\textbf{Sensitivity test on the scalar coefficient} $\lambda_2$ of $\mathcal{L}_\text{kd}$ term of the test time loss $\mathcal{L}_\text{TeSLA}$. We report, for each case, the average error rate over the 4 validation corruptions.}
    \label{fig:lmb2_ablation}
\end{figure}

\paragraph{Knowledge distillation coefficient.}
Finally, we report the sensitivity test results for the knowledge distillation weight $\lambda_2$. In Fig. \ref{fig:lmb2_ablation}, we show the classification adaptation performance of TeSLA on the CIFAR10-C and CIFAR100-C datasets is not very sensitive to the selection of $\lambda_2 \in \{0.1, 1, 10\}$.

\subsection{Ablations}
\begin{figure}
\centering
\begin{subfigure}{0.48\linewidth}
    \includegraphics[width=\linewidth]{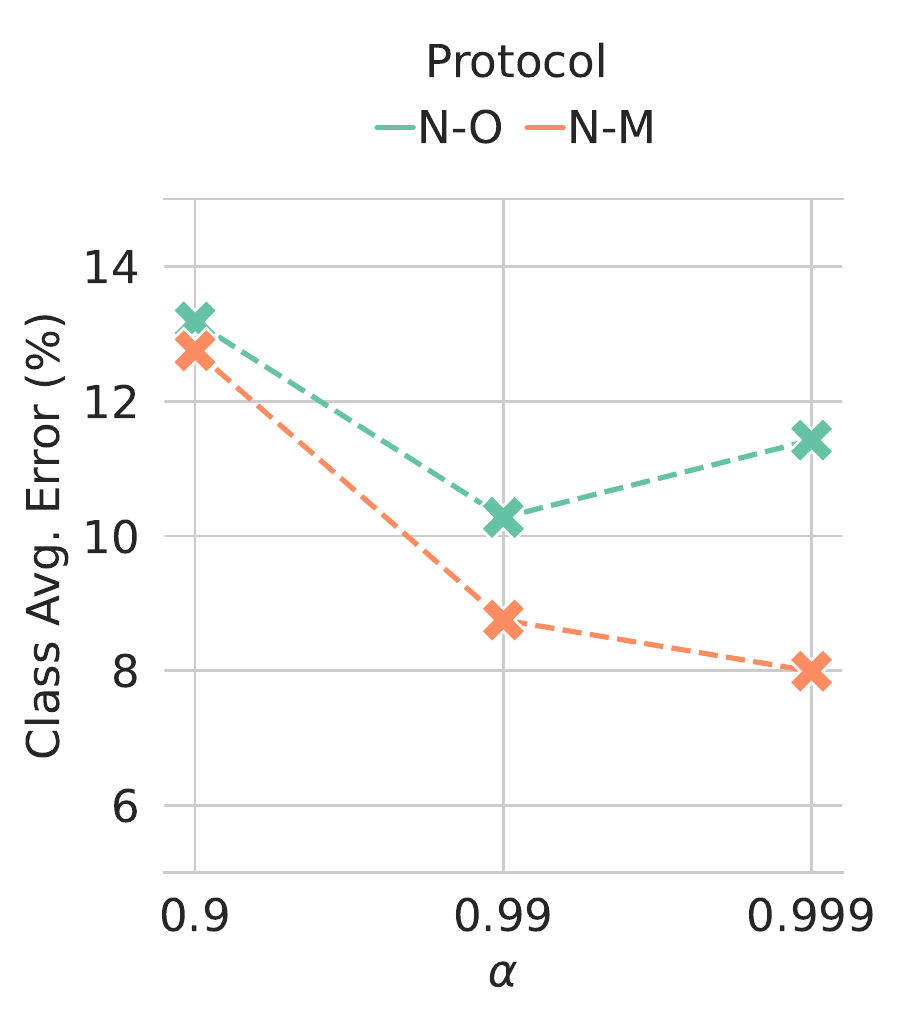}
    \caption{CIFAR10-C}
\end{subfigure}
\begin{subfigure}{0.48\linewidth}
    \includegraphics[width=\linewidth]{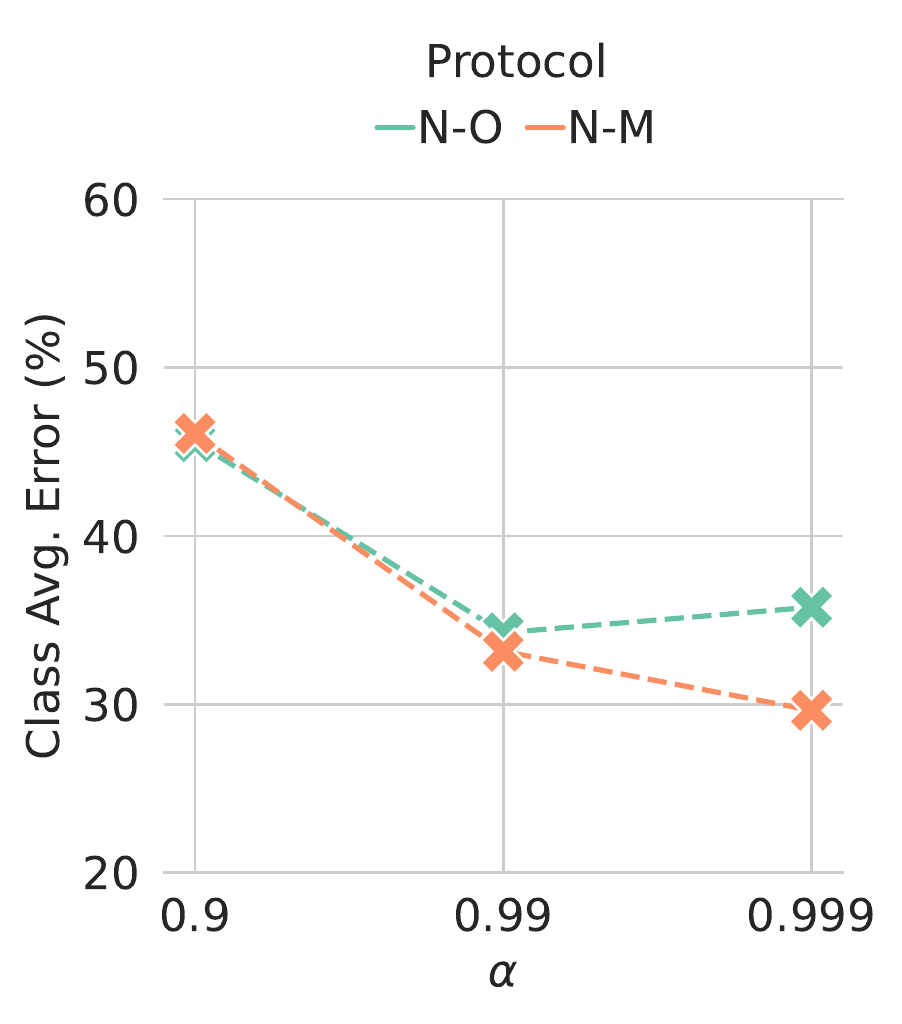}
    \caption{CIFAR100-C}
\end{subfigure}
    \caption{\textbf{Sensitivity test on the EMA coefficient of the teacher model} $\alpha$ on the (a) CIFAR10-C and (b) CIFAR100-C datasets. We report the average error rate over four corruptions for each dataset.}
    \label{fig:alpha_ablation_cifar10}
\end{figure}

\paragraph{EMA coefficient of teacher model.}
In Fig. \ref{fig:alpha_ablation_cifar10}, we show the effect of changing the EMA coefficient $\alpha$ used for updating the teacher model from the student model for the one-pass (O protocol) and multi-pass (M protocol) on the CIFAR10-C and CIFAR100-C datasets. We observe that for the multi-pass protocol (M), decreasing $\alpha$ leads to better performance, while for the one-pass protocol (O), optimal $\alpha$ depends on the number of test images observed in one epoch. If $\alpha$ is large (close to 1.0), the teacher is updated very slowly and thus requires more updates to reach better performance. Therefore, for a one-pass online evaluation, the accuracy decreases. On the other hand, if we set $\alpha$ to a minimal value, it results in unstable convergence.

\begin{figure}
 \centering
 \begin{subfigure}{0.48\linewidth}
    \includegraphics[width=\linewidth]{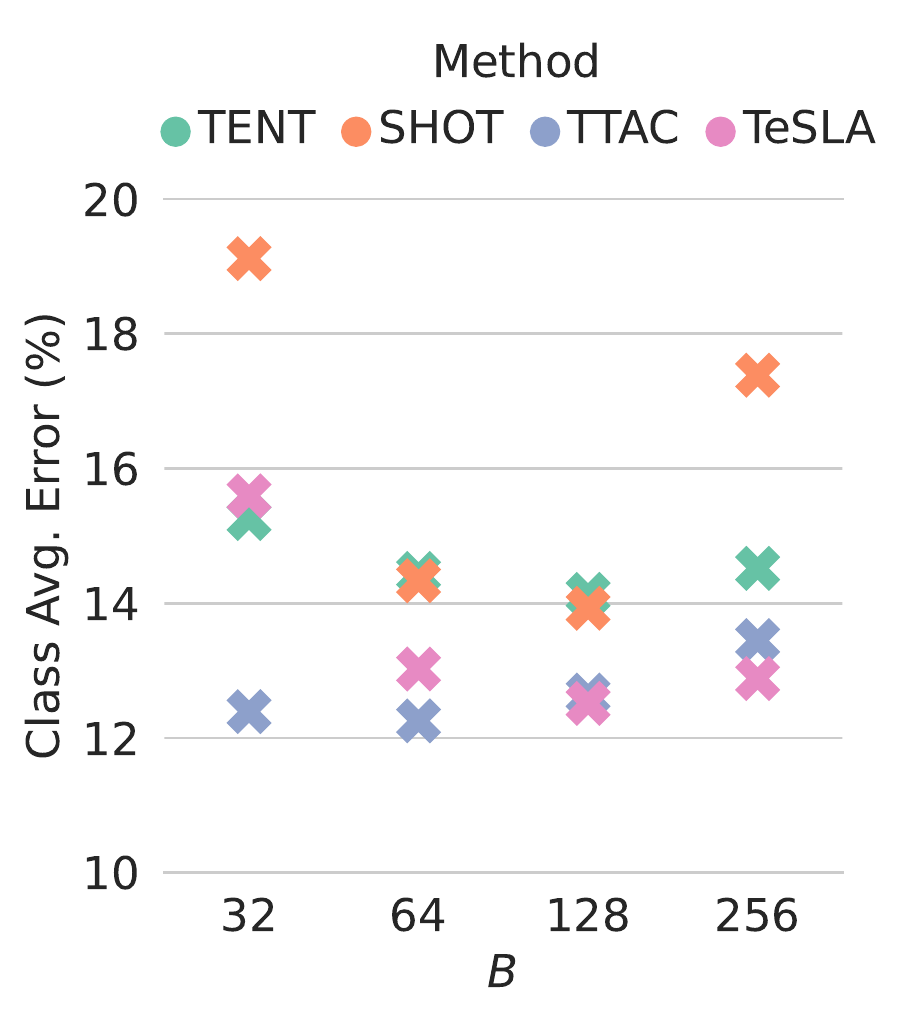}
    \caption{}
\end{subfigure}
\begin{subfigure}{0.48\linewidth}
    \includegraphics[width=\linewidth]{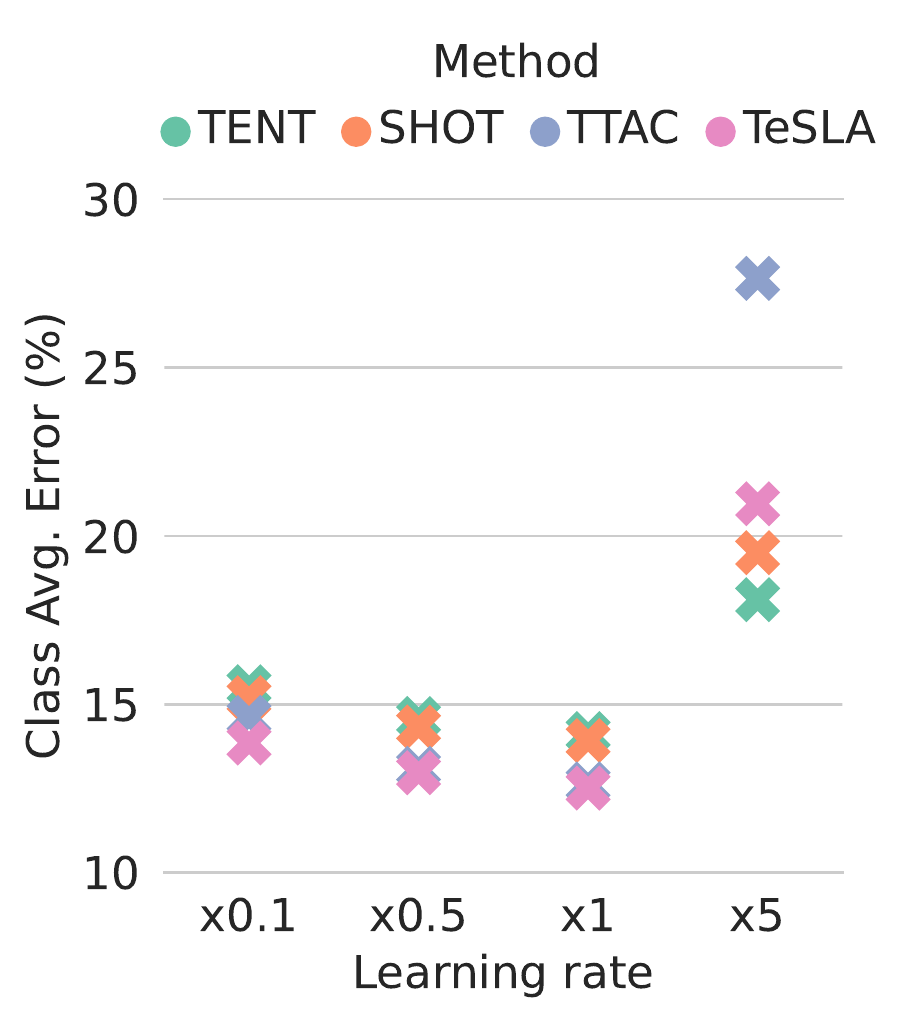}
    \caption{}
 \end{subfigure}
    \caption{\textbf{Ablation study} for the roles played by the (a) batch size $B$ and (b) the learning rate scale with respect to the default value. We report, for each baseline, the average error rate (in \%) over four validation corruption sets of the CIFAR10-C under the N-O protocol.}
    \label{fig:bs_lr}
\end{figure}

\paragraph{Batch size and learning rate.}
In Fig. \ref{fig:bs_lr}, we show the effect of batch size and learning rate on the proposed method TeSLA along with TENT\cite{wang2021tent}, SHOT\cite{liang2020we}, and TTAC\cite{su2022revisiting} on CIFAR10-C for N-O protocol. We observe that increasing batch size helps reduce test time error rates, and the model performs best with the same batch size used during source model training. Similarly, increasing the learning rate reduces the error rate until it becomes too large for unstable gradient model updates.

\section{Additional Qualitative Results}\label{appendix:qualitative_results}
\renewcommand\thefigure{\thesection.\arabic{figure}}
\renewcommand\thetable{\thesection.\arabic{table}}
\setcounter{figure}{0}
\setcounter{table}{0}
\begin{figure}[t]
    \centering
    \includegraphics[width=0.9\linewidth]{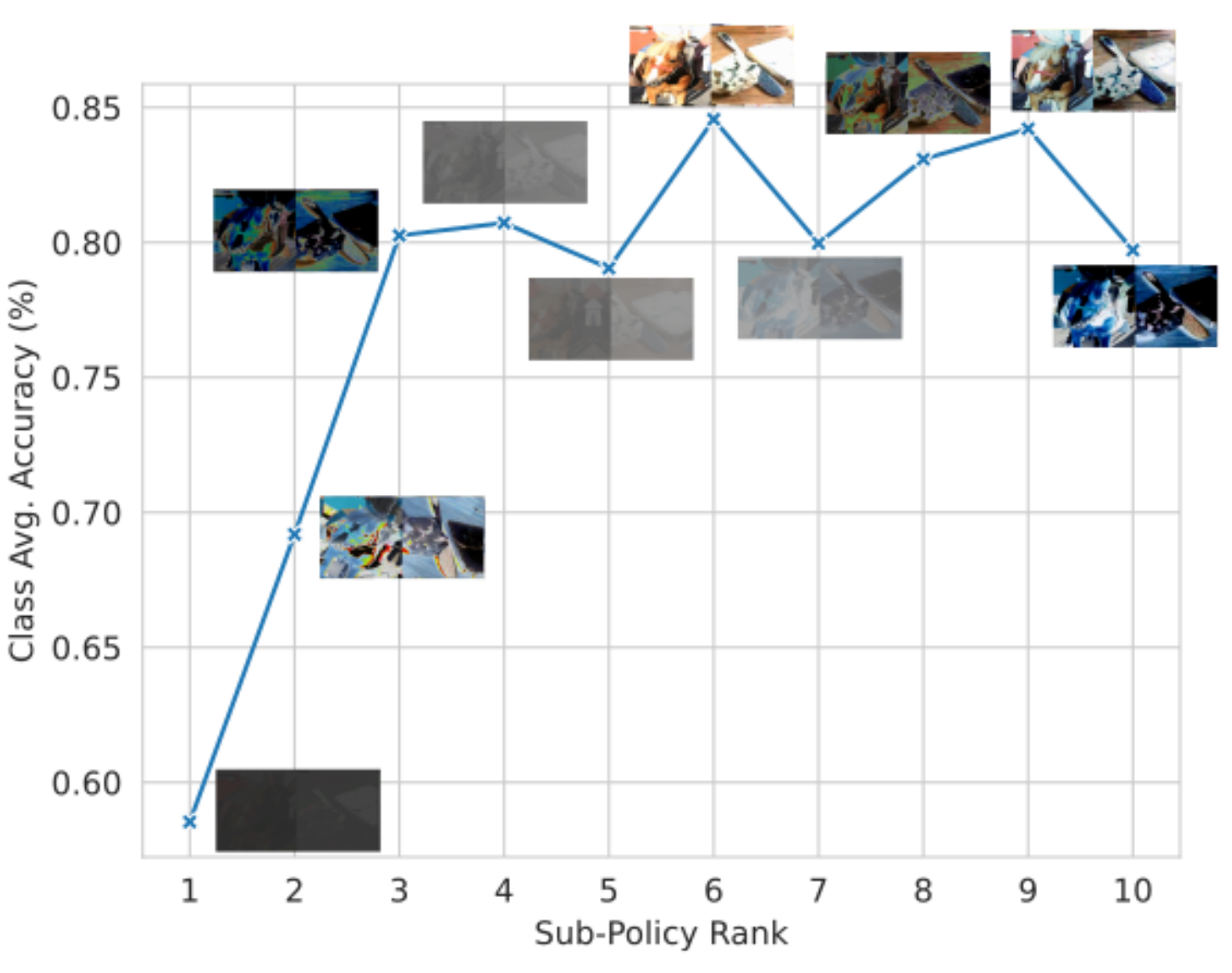}
    \caption{\textbf{Adversarial augmentation sanity check.} We report the per-class average accuracy (\%) of TeSLA's student model on the VisDA-C augmented by the ten most adversarial sub-policies optimized by our automatic augmentation module.}
    \label{fig:sanity_check}
\end{figure}
\begin{figure}
    \centering
    \includegraphics[width=\linewidth]{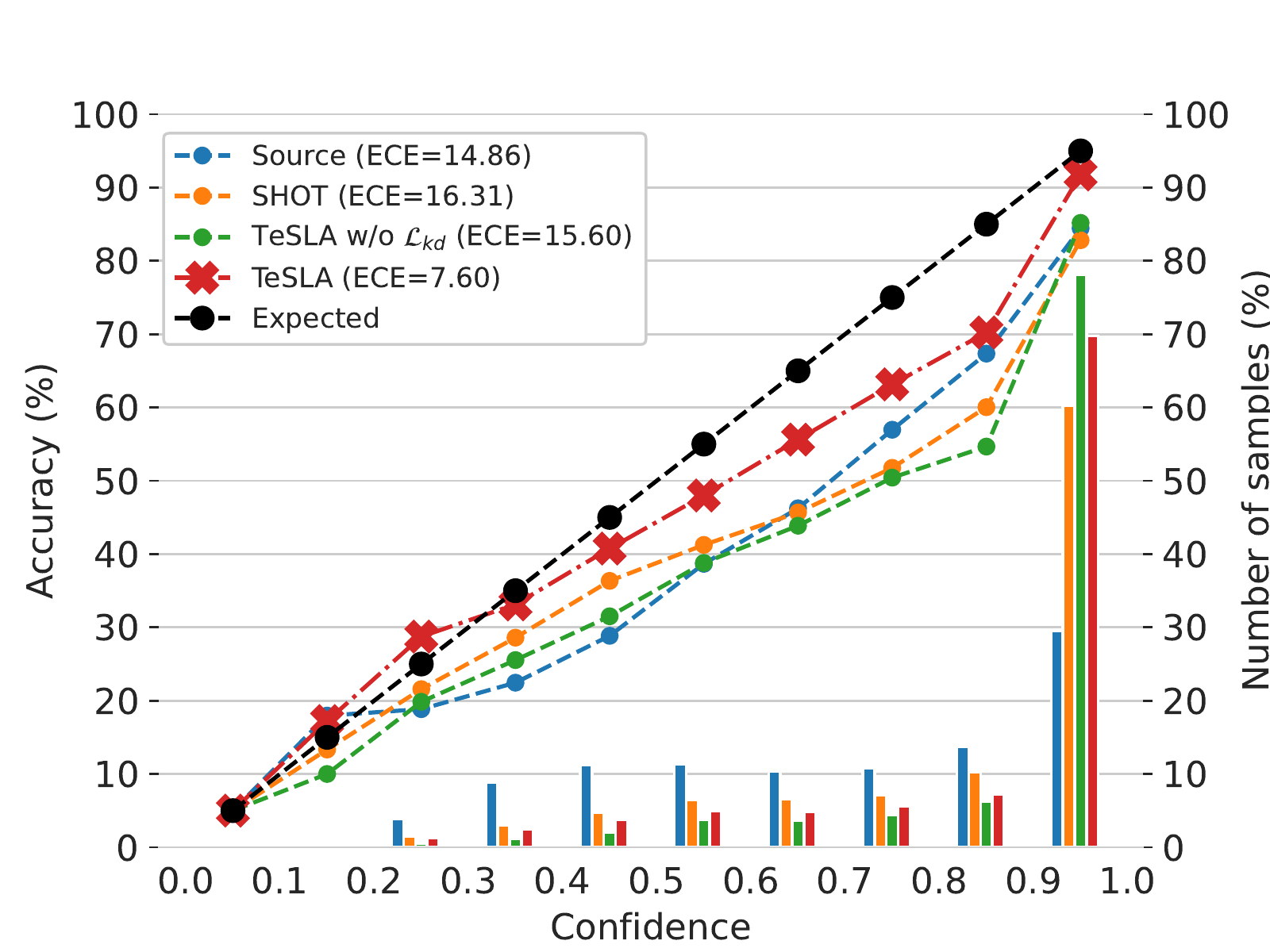}
    \caption{\textbf{Calibration performance comparison} of TeSLA (with and without adversarial augmentations) against other baselines via a reliability diagram on the VisDA-C dataset for N-O protocol.}
    \label{fig:calibration}
\end{figure}
\subsection{Sanity Check for Adversarial Augmentation}
To assess the adversarial effect of the proposed automatic augmentations, we conduct a sanity check for the optimized sub-policies. in particular, in Fig. \ref{fig:sanity_check}, we rank the sub-policies optimized by our automatic augmentation module on the VisDA-C for N-M protocol after one epoch by decreasing the order of sampling probability. Then, we evaluate the performance of the student model on the test-test images from VisDA-C that are augmented using the above sub-policies. We observe that reducing the hardness level of sub-policies, the more the student model is accurate in recognizing the images. This is supported by the Pearson correlation of 0.7 between the sub-policy rank and the accuracy $(p=0.02)$, demonstrating our module's capability to optimize and sample adversarial examples.

\begin{figure}[t]
    \centering
    \resizebox{\linewidth}{!}{
    \begin{subfigure}{0.15\linewidth}
        \includegraphics[width=\linewidth]{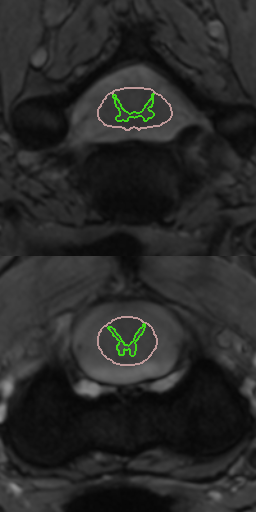}
        \caption{}
    \end{subfigure}
    \begin{subfigure}{0.15\linewidth}
        \includegraphics[width=\linewidth]{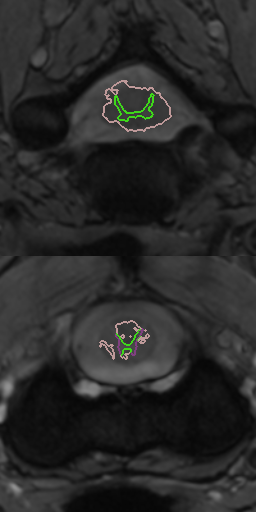}
        \caption{}
    \end{subfigure}
    \begin{subfigure}{0.15\linewidth}
        \includegraphics[width=\linewidth]{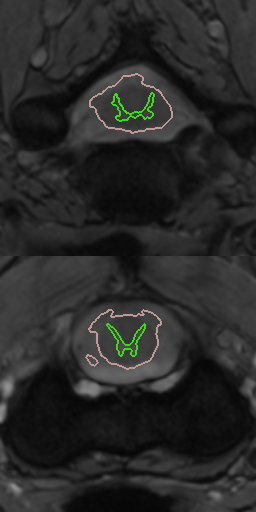}
        \caption{}
    \end{subfigure}
    \begin{subfigure}{0.15\linewidth}
        \includegraphics[width=\linewidth]{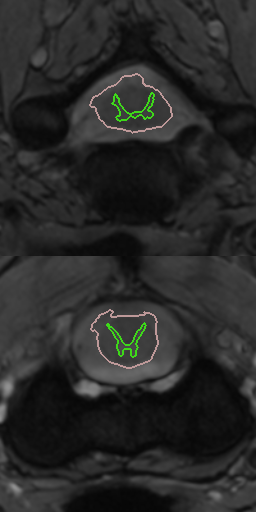}
        \caption{}
    \end{subfigure}
    \begin{subfigure}{0.15\linewidth}
        \includegraphics[width=\linewidth]{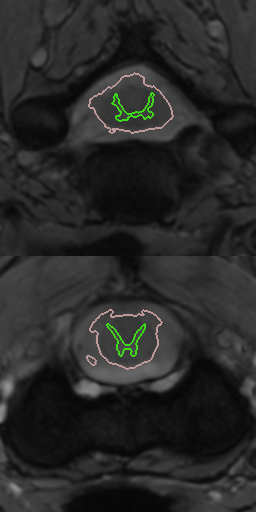}
        \caption{}
    \end{subfigure}
    \begin{subfigure}{0.15\linewidth}
        \includegraphics[width=\linewidth]{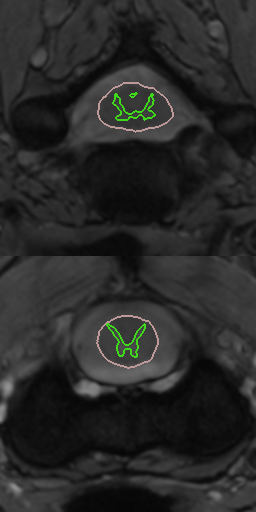}
        \caption{}
    \end{subfigure}}
    \caption{\textbf{Qualitative segmentation results} of test-time adaptation methods trained on \textbf{site 1} and tested on \textbf{site 3} of the spinal cord dataset. From left to right: (a) Ground Truth, (b) Source Model, (c) BN\cite{nado2020evaluating}, (d) TENT\cite{wang2021tent}, (e) PL, and (f) TeSLA, respectively.}
    \label{fig:spinal_qulatitative}
\end{figure}

\begin{figure}
    \centering
    \resizebox{\linewidth}{!}{
    \begin{subfigure}{0.15\linewidth}
        \includegraphics[width=\linewidth]{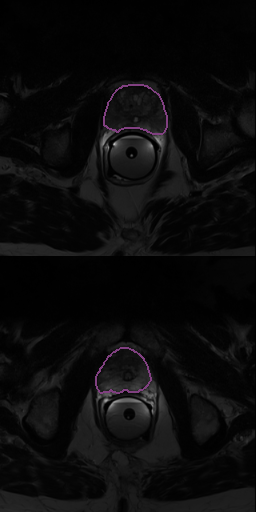}
        \caption{}
    \end{subfigure}
    \begin{subfigure}{0.15\linewidth}
        \includegraphics[width=\linewidth]{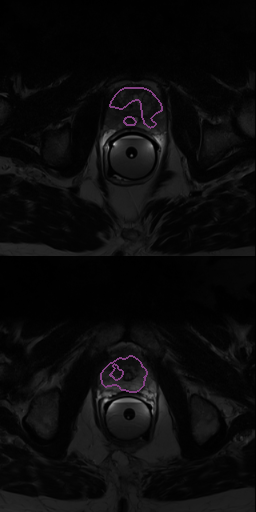}
        \caption{}
    \end{subfigure}
    \begin{subfigure}{0.15\linewidth}
        \includegraphics[width=\linewidth]{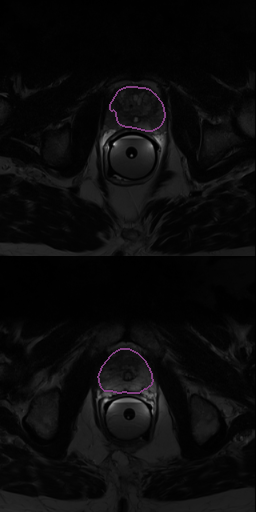}
        \caption{}
    \end{subfigure}
    \begin{subfigure}{0.15\linewidth}
        \includegraphics[width=\linewidth]{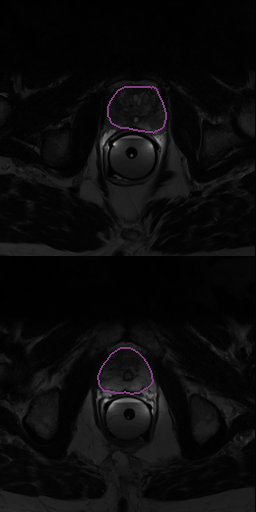}
        \caption{}
    \end{subfigure}
    \begin{subfigure}{0.15\linewidth}
        \includegraphics[width=\linewidth]{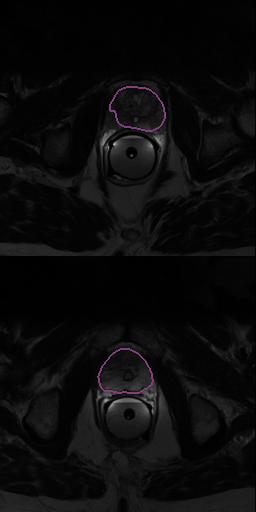}
        \caption{}
    \end{subfigure}
    \begin{subfigure}{0.15\linewidth}
        \includegraphics[width=\linewidth]{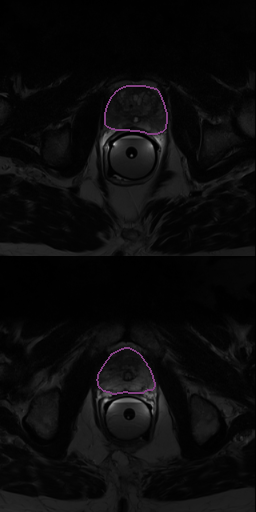}
        \caption{}
    \end{subfigure}}
    \caption{\textbf{Qualitative segmentation results} of test-time adaptation method trained on \textbf{site A} and \textbf{site B} and tested on \textbf{site F} of the prostate dataset. From left to right: (a) Ground Truth, (b) Source Model, (c) BN\cite{nado2020evaluating}, (d) TENT\cite{wang2021tent}, (e) PL, and (f) TeSLA, respectively.}
    \label{fig:prostate_qulatitative}
\end{figure}

\subsection{Uncertainty Evaluation}
We evaluate the model reliability on the ViSDA-C classification adaptation task (N-O protocol). In Fig. \ref{fig:calibration}, we show the reliability diagram (dividing the probability range [0, 1.0] into ten bins) and report the expected calibration error (ECE) \cite{niculescu2005predicting} for the Source model without adaptation, SHOT \cite{liang2020we}, and TeSLA with and without adversarial augmentations. The proposed TeSLA gives the lowest calibration error with an 8.71\% improvement over SHOT. It is interesting to observe that the ECE of TeSLA without adversarial augmentations is on-par with the SHOT method. The adversarial augmentation module improves the TeSLA's ECE by 8\%, which shows the benefit of test-time adversarial augmentation on the model's reliability.

\subsection{Qualitative Segmentation Results}
In Fig. \ref{fig:spinal_qulatitative} and Fig. \ref{fig:prostate_qulatitative}, we show the qualitative segmentation results of TeSLA for test-time adaptation on the spinal cord and prostate MRI datasets and compare it with TENT\cite{wang2021tent}, BN\cite{nado2020evaluating}, and Pseudo Labeling (PL). Compared to other baselines, TeSLA outputs more accurate segmentation results closer to the provided ground truth.

\end{document}